\documentclass{article}

\usepackage[preprint]{neurips_2023}

\usepackage[utf8]{inputenc} 
\usepackage[T1]{fontenc}    
\usepackage{url}            
\usepackage{booktabs}       
\usepackage{amsfonts}       
\usepackage{nicefrac}       
\usepackage{microtype}      
\usepackage{xcolor}         
\usepackage{makecell}
\usepackage{multirow}
\usepackage{caption}
\usepackage{graphicx}
\usepackage{amsmath}
\usepackage{subcaption}
\usepackage{wrapfig}

\usepackage{natbib}
\setcitestyle{numbers,square}
\usepackage[colorlinks,linkcolor=black,anchorcolor=black,citecolor=black]{hyperref}
\usepackage{amssymb}
\newcommand{\KMEquation}[1]{Eq.~(\ref{#1})}
\newcommand{\KMFigure}[1]{Figure.~\ref{#1}}
\newcommand{\KMTable}[1]{Table~\ref{#1}}
\newcommand{\KMSec}[1]{\S~\ref{#1}}
\newcommand{\eg}{\textit{e.g.,} }
\newcommand{\ie}{\textit{i.e.,} }

\newcommand{\redContent}[1]{{\color{red}#1}}
\newcommand{\blueContent}[1]{{\color{blue}#1}}

\definecolor{seagreen}{rgb}{0.18, 0.55, 0.34}
\definecolor{royalpurple}{rgb}{0.47,0.32,0.66}
\definecolor{brown(traditional)}{rgb}{0.59, 0.29, 0.0}
\definecolor{blue(traditional)}{rgb}{0.15, 0.29, 0.50}
\definecolor{azure}{rgb}{0, 0.5, 1}
\newcommand{\zqh}[1]{{\color{black}#1}}
\usepackage{amsmath,amssymb,amsfonts,mathtools,amsthm}

\title{Dissecting Arbitrary-scale Super-resolution Capability from Pre-trained Diffusion Generative Models}

\author{%
Ruibin Li, Qihua Zhou, Song Guo, Jie Zhang, Jingcai Guo, \\ 
\textbf{Xinyang Jiang, Yifei Shen, Zhenhua Han}
}

\begin{document}

\maketitle

\begin{abstract}

Diffusion-based Generative Models (DGMs) have achieved unparalleled performance in synthesizing high-quality visual content, opening up the opportunity to improve image super-resolution (SR) tasks. Recent solutions for these tasks often train architecture-specific DGMs from scratch, or require iterative fine-tuning and distillation on pre-trained DGMs, both of which take considerable time and hardware investments. More seriously, since the DGMs are established with a discrete pre-defined upsampling scale, they cannot well match the emerging requirements of arbitrary-scale super-resolution (ASSR), where a unified model adapts to arbitrary upsampling scales, instead of preparing a series of distinct models for each case. These limitations beg an intriguing question: \textit{can we identify the ASSR capability of existing pre-trained DGMs without the need for distillation or fine-tuning?} In this paper, we take a step towards resolving this matter by proposing Diff-SR, a first ASSR attempt based solely on pre-trained DGMs, without additional training efforts. It is motivated by an exciting finding that a simple methodology, which first injects a specific amount of noise into the low-resolution images before invoking a DGM's backward diffusion process, outperforms current leading solutions. The key insight is determining a suitable amount of noise to inject, \ie small amounts lead to poor low-level fidelity, while over-large amounts degrade the high-level signature. Through a finely-grained theoretical analysis, we propose the \textit{Perceptual Recoverable Field} (PRF), a metric that achieves the optimal trade-off between these two factors. Extensive experiments verify the effectiveness, flexibility, and adaptability of Diff-SR, demonstrating superior performance to state-of-the-art solutions under diverse ASSR environments.

\end{abstract}

\section{Introduction}

\zqh{Over the last decade, the deep neural models (\eg EDSR \cite{EDSR, EDSR2}, ESRGAN \cite{ESRGAN, ESRGAN2}) have significantly promoted the development of super-resolution (SR) techniques. 
However, it is still challenging for them to match the emerging requirements of arbitrary-scale super-resolution (ASSR) tasks \cite{ASSR}. The primary target of ASSR is to provide a unified model for arbitrary upsampling scales, instead of training a series of distinct models for each case. 
Traditional SR models are often customized to a specific integer scale setting (\eg $2\times$).
Thus, when dealing with a larger scale (\eg $4\times$) during inference, a natural way is to cascade a $2\times$ SR model twice or train a new $4\times$ SR model from scratch \cite{SISR}. Unfortunately, previous work has shown that this method often suffers from losing high-fidelity details \cite{SISR2}.} Besides, due to the fixed network architecture, it is hard to adapt the model to non-integer scale SR tasks, \eg $2.7\times$. Observing these issues, the recent LIIF \cite{LIIF} and its variations \cite{LIIF2,LIIF3} try to learn a continuous function representing high-resolution images. However, their upscaled images still suffer from unacceptable structural distortion and fidelity loss.

\zqh{Recently, Diffusion-based Generative Models (DGMs) \cite{DBLP:conf/icml/Sohl-DicksteinW15}, have achieved remarkable success in synthesizing high-quality visual content \cite{SD,DallE,DbeatGAN, VideoDFM3,DFMlongV, DDGM,SDE}. 
Due to this unique strength \cite{DBLP:conf/nips/DhariwalN21}, DGMs have opened up the opportunity to handle image SR tasks \cite{DBLP:journals/corr/abs-2104-07636}.
Although DGMs can be implemented with large freedom, in this paper, we will exclusively demonstrate our methodology based on Denoising Diffusion Probabilistic Modeling (DDPM) \cite{DBLP:conf/nips/HoJA20}, which is a pertinent case belonging to the DGM family.}
Generally, DDPM is inspired by non-equilibrium thermodynamics \cite{DBLP:conf/icml/Sohl-DicksteinW15}. It defines a Markov chain \cite{Markovchain} of diffusion steps by gradually adding Gaussian noise into data, and learns to reverse the diffusion process to reconstruct data samples from the noise \cite{weng2021diffusion}. For image synthesis, DGMs start from a generative seed based on Gaussian noise, and then iteratively denoise it to obtain a clear image with high perceptual quality. 
\zqh{Despite the benefit, most recent solutions often train architecture-specific DGMs from scratch, or require the efforts of fine-tuning \cite{finetuning,finetuning2,finetuning3} and distillation \cite{Distillation1,Distillation2,Distillation3}, both of which take considerable time and hardware investments. More seriously, the DGMs are established based on a discrete pre-defined upsampling scale, limiting their ASSR performance when adapting to a different scale during inference.}

\zqh{These limitations raise an interesting question -- \textit{can we identify the ASSR capability from an existing pre-trained DGM without additional efforts of fine-tuning or distillation?}
Surprisingly, we find a simple but effective methodology to resolve this matter, \ie injecting a specific amount of noise into the low-resolution (LR) image before invoking a DGM's backward diffusion process.
Our motivation is that since the DGMs can generate high-quality visual content and the target of ASSR is also recovering LR images by generating visual details, we can utilize this property to control a DGM's pipeline and adapt it to the ASSR environments. 
The key here is to determine a suitable amount of noise to inject. We develop insights that the injected noise affects the ASSR capability in both low-level fidelity measure and high-level signature of generated content. 
From this perspective, we prove the feasibility of this methodology in theory and deduce the \textit{Perceptual Recoverable Field} (PRF). This key concept indicates how much noise could be injected to guarantee a good recovery quality for different upsampling scales. 
Based on these theoretical fundamentals, we deeply analyze the rationale of DGM's image ASSR capacity and give a mathematical analysis of the suitable noise injection level to obtain the desired recovery quality.}

\zqh{We implement our methodology as Diff-SR, a first ASSR attempt based on a single pre-trained DGMs solely.
For real-world ASSR deployment, Diff-SR just involves the inference process of DGMs. Diff-SR injects a specific amount of noise into the LR images and provides a unified generative starting point for visual details recovery. By invoking the reverse diffusion process, Diff-SR can restore the noisy LR images into the high-resolution version, with similar perceptual quality as the ground truth. 
Evaluations show that our Diff-SR outperforms state-of-the-art solutions with better FID, PSNR and SSIM scores.}
\zqh{
Overall, the key contributions of our work are as follows:

\begin{itemize}
\item \textbf{Pioneering Methodology.} We are the first attempt to identify a DGM's ASSR capability without distillation or fine-tuning. Excitingly, we find a simple but effective methodology for this matter that injects a specific amount of noise into the low-resolution image before invoking a DGM's backward diffusion process.

\item \textbf{Theoretical Guarantee.} We establish theoretical analysis to understand our methodology and quantify the ASSR capacity by deducing the key concept called \textit{Perceptual Recoverable Field} (PRF). Based on these fundamentals, we provide mathematical analysis to guarantee the noise injection strength for handling different ASSR tasks. 

\item \textbf{Efficient Implementation.} We implement our methodology as Diff-SR, a novel ASSR solution based solely on a single pre-trained DGM. Evaluations based on real-world settings verify the superiority of Diff-SR over the current leading solutions.
\end{itemize}
}

\section{Background}

\subsection{Forward Diffusion Process}

For diffusion model, it models the whole forward process as a Markov chain. We add a small amount of Gaussian noise for each step in the chain to convert the original image to a low-quality version. Each step is modeled by a Gaussian distribution where the noise strength is controlled by a variance schedule $\{\beta_t \in (0,1)\}_{t=1}^T$. Thus, we can be described a step as follows:

\begin{equation}
q(\mathbf{x}_t \vert \mathbf{x}_{t-1}) = \mathcal{N}(\mathbf{x}_t; \sqrt{1 - \beta_t} \mathbf{x}_{t-1}, \beta_t\mathbf{I}) \quad
q(\mathbf{x}_{1:T} \vert \mathbf{x}_0) = \prod^T_{t=1} q(\mathbf{x}_t \vert \mathbf{x}_{t-1}).
\label{Equation:forward}
\end{equation}

By utilizing the property of Gaussian distribution, we can formulate the distribution of $x_t$ given $x_0$ as:

\begin{equation}
q(\mathbf{x}_t \vert \mathbf{x}_0) = \mathcal{N}(\mathbf{x}_t; \sqrt{\bar{\alpha}_t} \mathbf{x}_0, (1 - \bar{\alpha}_t)\mathbf{I}),
\label{Equation:forward_t}
\end{equation}
where $\alpha_t=1-\beta_t$ and $\bar{\alpha}_t = \prod_{i=1}^t \alpha_i$. Usually, the noise strength increases along with time $\beta_1<\beta_2<...<\beta_T$. 
Therefore, we have $\bar{\alpha}_1>\bar{\alpha}_2>...>\bar{\alpha}_T$. With this condition distribution, we can derive the posterior distribution of $x_{t-1}$ conditioned by $x_t,x_0$ as:

\begin{equation}
\begin{aligned}
q(\mathbf{x}_{t-1} \vert \mathbf{x}_t, \mathbf{x}_0) &= \mathcal{N}(\mathbf{x}_{t-1}; \tilde{\boldsymbol{\mu}}(\mathbf{x}_t, \mathbf{x}_0), {\tilde{\beta}_t} \mathbf{I}). 
\end{aligned}
\label{Equation:backward}
\end{equation}

\subsection{Backward Diffusion Process}
The target of a DGM is to maximize the likelihood probability $p_{\theta}(x_0)$. 
Based on the negative log-likelihood theorem, we can maximize $p_{\theta}(x_0)$ and optimize the variational upper bound by introducing a KL-divergence term \cite{KL2,KLdiver}. According to the preliminary formulation mentioned by \KMEquation{Equation:forward_t} and \KMEquation{Equation:backward}, we can deduce the variational lower bound $L_{VLB}$ as:
\vspace{-2pt}
\begin{equation}
\begin{aligned}
- \log p_\theta(\mathbf{x}_0) 
&\leq - \log p_\theta(\mathbf{x}_0) + D_\text{KL}(q(\mathbf{x}_{1:T}\vert\mathbf{x}_0) \| p_\theta(\mathbf{x}_{1:T}\vert\mathbf{x}_0) ) \\
&= \underbrace{L_T}_{\text{Forward error}} + \underbrace{L_{T-1} + \dots + L_0}_{\text{Backward error}} \\
&= L_\text{VLB}, 
\label{Equation:L_VLB}
\end{aligned}
\end{equation}
where: 
\begin{equation}
\begin{aligned}
L_T &= D_\text{KL}(q(\mathbf{x}_T \vert \mathbf{x}_0) \parallel p_\theta(\mathbf{x}_T)) = C_T, L_0 = - \log p_\theta(\mathbf{x}_0 \vert \mathbf{x}_1).\\
L_t &= \mathbb{E}_{\mathbf{x}_0, \boldsymbol{\epsilon}} \Big[\frac{ (1 - \alpha_t)^2 }{2 \alpha_t (1 - \bar{\alpha}_t) \| \boldsymbol{\Sigma}_t \|^2_2} \|\boldsymbol{\epsilon}_t - \boldsymbol{\epsilon}_\theta(\textbf{x}_t, t)\|^2 \Big]  \\
\end{aligned}
\end{equation}

Note that $L_T$ reflects the forward error since it models the difference between the forward distribution $q(\mathbf{x}_T|\mathbf{x}_0)$ and the distribution of the neural network output. The other parts are the backward error, which measures the difference between the true backward distribution $q(\mathbf{x}_{t-1}|\mathbf{x}_t,\mathbf{x}_0)$.

\section{Method}

\subsection{Problem Formulation}

The aim of ASSR \cite{SR1,SR2} is to enhance multi-scale blur low-resolution images $\hat{\mathbf{x}}$ to get clear high-resolution images $\mathbf{x}$ with only one model, which is one kind of image-to-image translation task. The key formulation is $\min \mathcal{L} = \parallel p_\theta(\hat{\mathbf{x}}),\mathbf{x} \parallel^2$, where $\theta$ represents the SR model and $\hat{\mathbf{x}}$ can be any down scale version of original images. The key is to use one model to minimize the difference between the recovered image and the original image.

\subsection{Observations from DGM Denoising Process}

\begin{figure}[htp]
  \centering
    \includegraphics[width=0.95\linewidth]{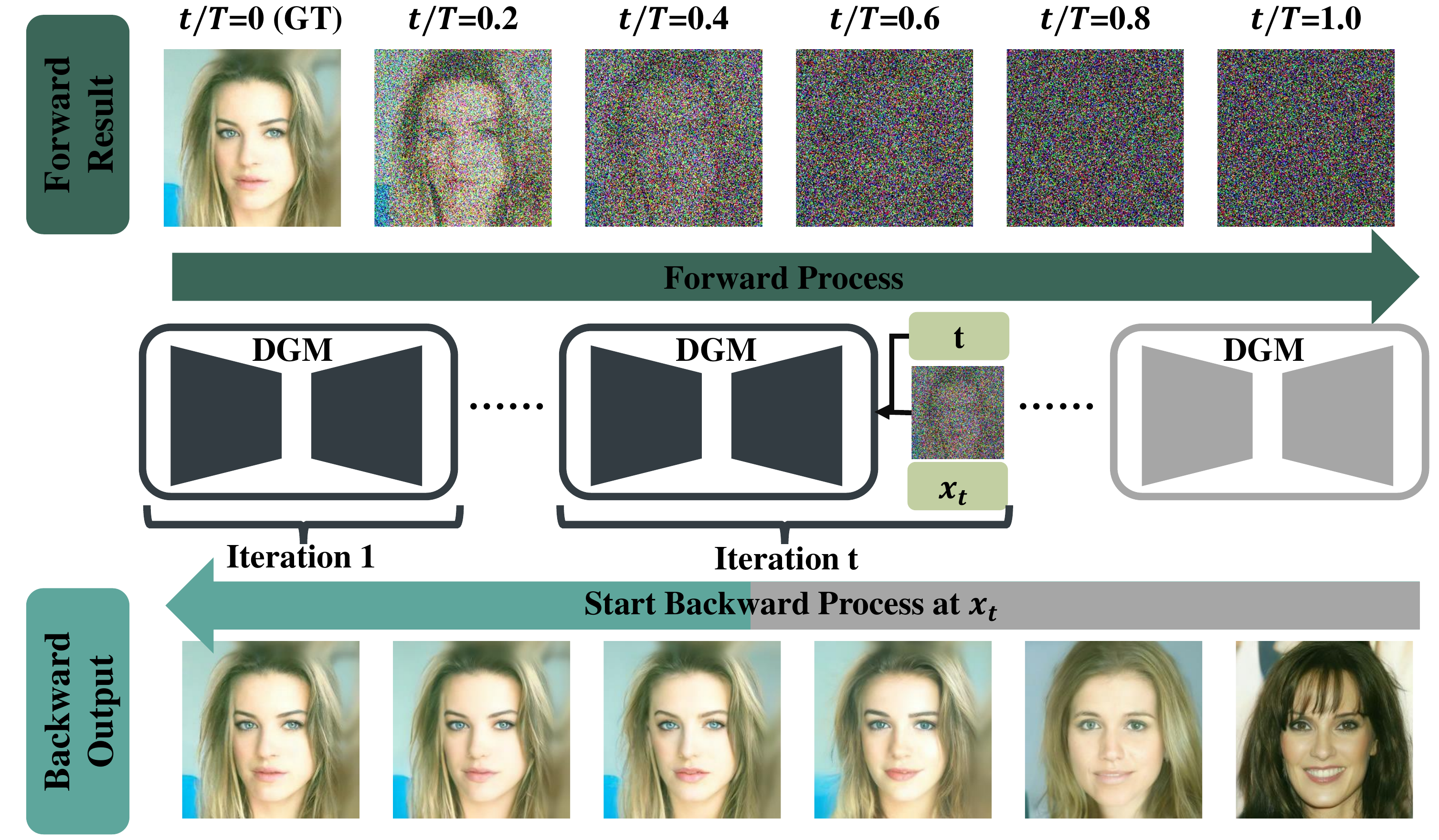}
  \caption{Illustration of DGM denoising process.}
  \label{Figure:reserve}
\vspace{-15pt}
\end{figure}

As illustrated in \KMFigure{Figure:reserve}, when initiating the backward process at timestep $t$ rather than $T$, a distinct output can still be obtained from the noisy input image. In this context, we denote $\bar{L}_t$ as the sampling error when commencing at $\mathbf{x}_t$, where $\mathbf{x}_t$ represents the outcome after injecting $t$ forward noise steps to the original image. Based on this, the following lemma is established:

\noindent\textbf{Lemma 1.} \textit{The error between the output image and the GT $\mathbf{x}_0$ can be formulated as:}
\begin{equation}
\begin{aligned}
\bar{L}_t 
&= C_t + \sum_{i=1}^{t-1} \Big[\frac{ (1 - \alpha_i)^2 }{2 \alpha_i (1 - \bar{\alpha}_i) \| \boldsymbol{\Sigma}_i \|^2_2} E_i \Big] + L_0 ,\\
\end{aligned}
\label{Equation:domain}
\end{equation}

where $E_i = \|\boldsymbol{\epsilon}_i - \boldsymbol{\epsilon}_\theta(\sqrt{\bar{\alpha}_i}\mathbf{x}_0 + \sqrt{1 - \bar{\alpha}_i}\boldsymbol{\epsilon}_i, i)\|^2$. Note that $\bar{L}_t$ models the error between the output image and the original image. $E_i$ is the training loss of the neural network and can be estimated empirically once the network converges. $E_i$ is actually the loss function of the diffusion model. When the neural network converges, this value can be regarded as a constant $E_0$. $\boldsymbol{\Sigma}_i$ is the variance of the reverse process which changes along with time $i$. As there are many research about how to speed up the sampling process of difussion model like DDIM \cite{DBLP:conf/iclr/SongME21}, this parameter may differ in the different sampler. For the basic DDPM \cite{DBLP:conf/nips/HoJA20}, $\boldsymbol{\Sigma}_t^{DDPM} = \tilde{\beta}_i \mathbf{I} $, for DDIM, 
$\boldsymbol{\Sigma}_t^{DDIM} = \frac{1 - \bar{\alpha}_{i-1}}{1 - \bar{\alpha}_i} \cdot \frac{1 - \bar{\alpha}_{i}}{\bar{\alpha}_{i-1}} \mathbf{I}$. 
Detailed proof can be found in the supplementary material of \S \ref{appendix:lemma1}.

\begin{figure}[htp]
  \centering
  \begin{subfigure}{0.49\linewidth}
    \includegraphics[width=\textwidth]{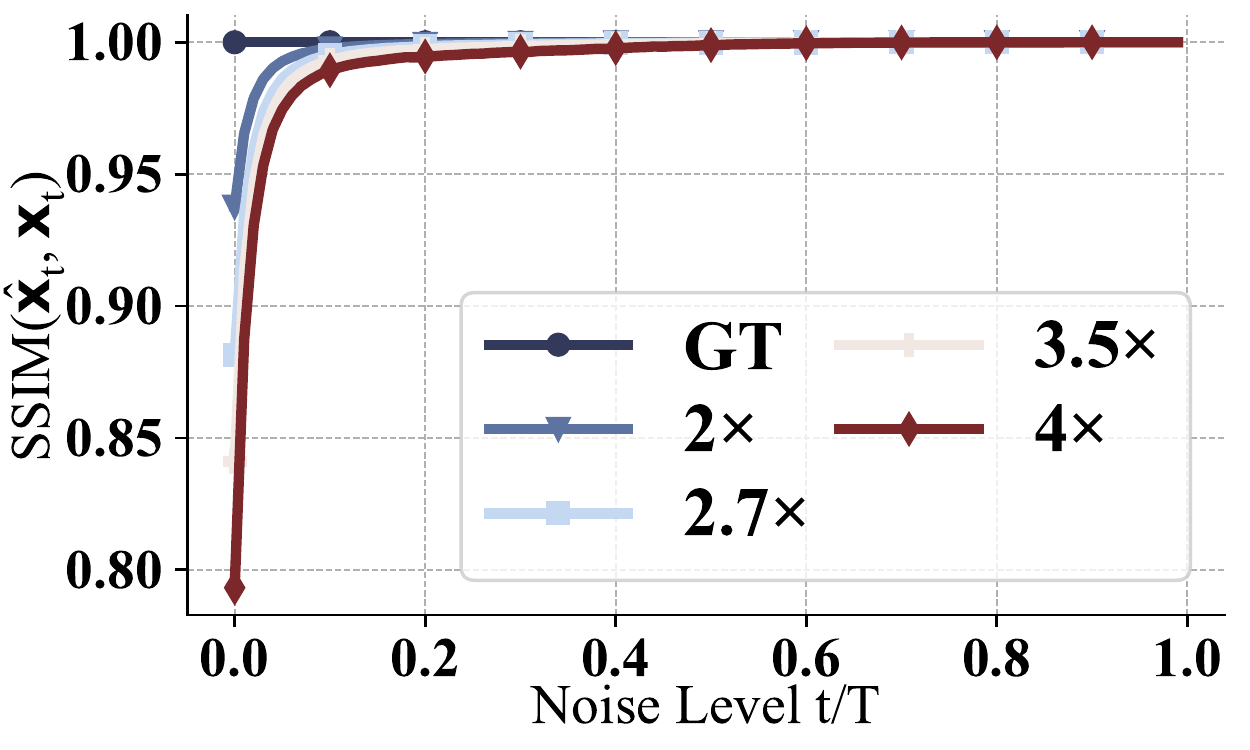}
    \caption{Similarity after injecting noise into LR images.\label{Figure:UniformDFM_similarity}}
  \end{subfigure}
  \begin{subfigure}{0.49\linewidth}
    \includegraphics[width=\textwidth]{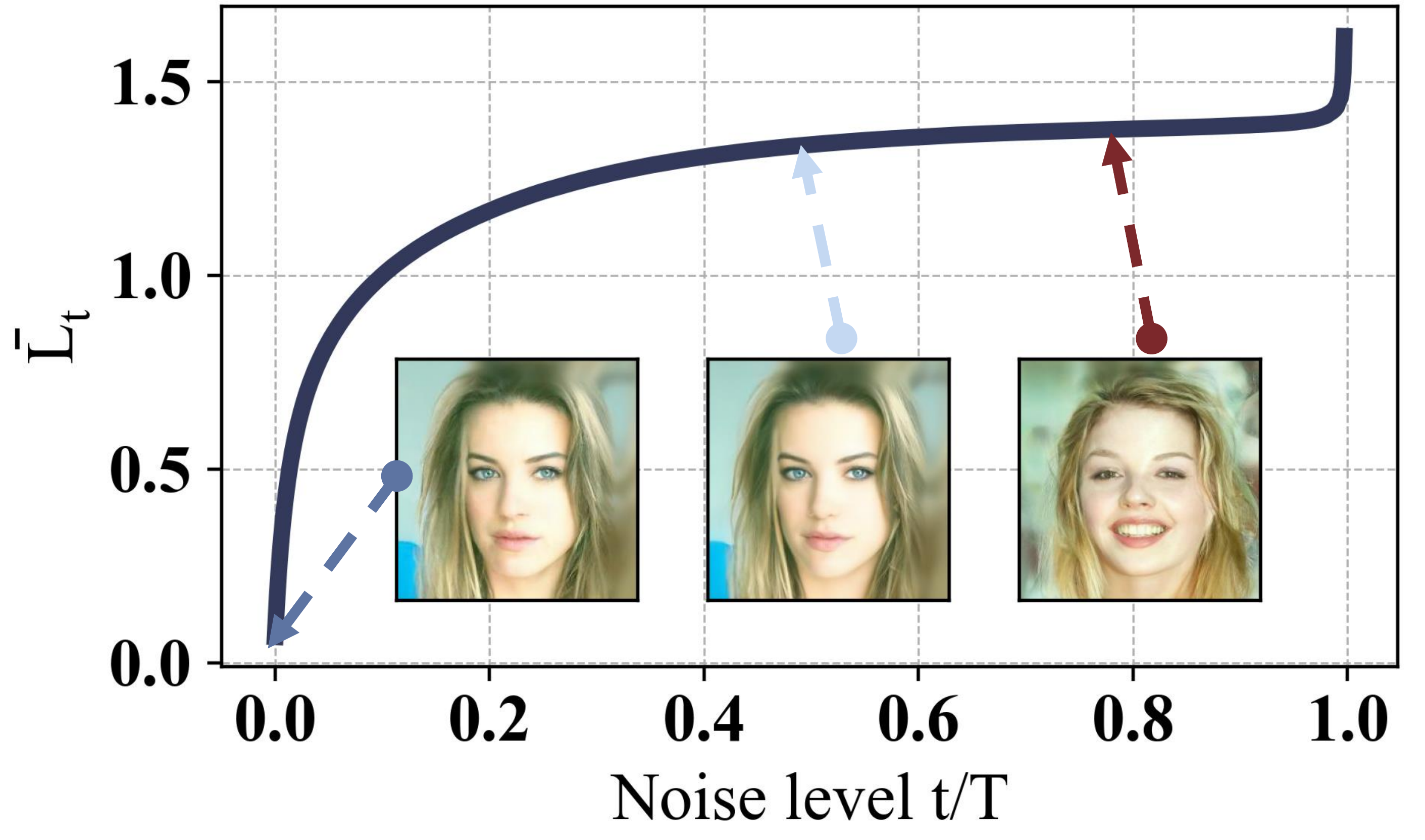}
    \caption{Recovered images under different noise levels.\label{Figure:sum_error}}
  \end{subfigure}%
  \caption{By injecting a specific amount of noise into the LR images, it is possible to recover the LR images to the HR version. However, the final perceptual quality of the recovered HR images may change with the injected noise level, which is controlled by step $t$.}
  \vspace{-15pt}
\end{figure}

Our first key observation is that this error can reflect the reversibility of the noisy image back to its original image. As depicted in \KMFigure{Figure:sum_error}, the behavior of $\bar{L}_t$ manifests a pronounced surge initially, spanning approximately 100 steps. During this phase, we note that the reconstructed image undergoes negligible changes in comparison to the original image. Subsequently, a gradual and smoother transition occurs up to around 600 steps. Throughout this interval, the reconstructed image largely retains the essential features of the original version, albeit with slight deviations in certain details. In the final phase, the reconstructed image diverges significantly from the original rendition. This intriguing phenomenon serves as a catalyst for further exploration, prompting us to investigate the potential applications of this characteristic.

\subsection{Potential of Dissecting ASSR Capability}

\begin{figure}[htp]
  \centering
    \includegraphics[width=0.99\linewidth]{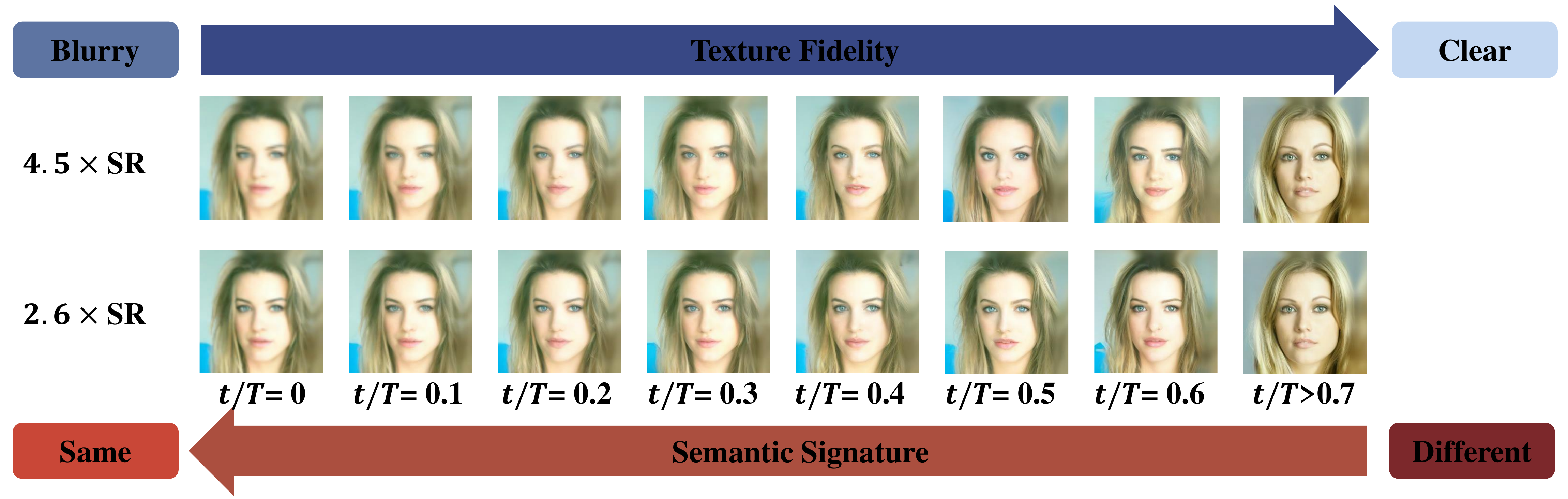}
  \caption{Image SR results of a pre-train DGM with different steps of noise injection. Take both $4.5 \times$ and $2.6 \times$ upsampling scales as the example, we inject different $t$ steps of noise into the input LR image, and then conduct the DGM's reverse process from this step to generate the HR version. We can see that the step numbers impact both texture fidelity and semantic signature, which are with opposite proportional relations to step $t$. (\textbf{Zoom-in for best view})}
  \label{Figure:compare_celep}
\vspace{-15pt}
\end{figure}

Then, inspired by our preliminary experiment, we found that after injecting some noise to low-resolution images $\hat{\mathbf{x}}$, the sampling results $\hat{\mathbf{x}}_t$ show high similarity with the high-resolution images $\hat{\mathbf{x}}$ as shown in \KMFigure{Figure:UniformDFM_similarity}. This adding noise process is the sampling process from the forward distribution $\mathbf{x}_t\sim q(\mathbf{x}_t \vert \mathbf{x}_0)$. The high similarity is because their distributions become similar. The distribution similarity can be measured by KL-divergence $D_{\text{KL}}(q(\mathbf{x}_t \vert \mathbf{x}_0) \parallel q(\hat{\mathbf{x}}_t \vert \hat{\mathbf{x}}_0) )$ where we have deeper analysis in \KMSec{section:noise_injection}. This inspires us to explore whether we can recover the high-resolution image with these noised low-resolution images, so we try to use different resolutions to conduct reserves steps after injecting a certain amount of noise into them. The source resolution is $256\times256$, we convert the source image to different versions, and the result is demonstrated in the \KMFigure{Figure:compare_celep}. According to the experiment, we can easily recover the original image when the resolution is near the original resolution (\ie downsampling the original image to $2.6\times$ scale), just inject $20\%$ noise into the degraded image, then we can recover the low-resolution image. However, as the resolution decreases, we should inject more noise, as shown in the $4.5\times$ scale downsampling, when we add $20\%$ noise and start the reverse process at this point, it ends up returning a blurry picture just as the input data, So we need to add more noise to the input data, then it can return a more clear picture that maintains the majority of the source image. Therefore there is a tradeoff which is if we add too much noise, it will finally return totally different images.

\subsection{Analysis of Recovery Error}
\label{section:noise_injection}

In this section, we analyze how to control the amount of noise injected and explain why it works. Note that $p_{\theta}$ is a well pretrained Diffusion model. By just injecting some noise into the low-resolution images and changing the start point of the reverse diffusion process, we can now use $p_{\theta}$ to conduct super-resolution tasks.

With a little abuse of symbols, we use $\mathbf{x}_{t},\hat{\mathbf{x}}_{t}$ to represent the result after we inject $t$ steps noise to the original high-resolution images and low-resolution images, respectively. $\bar{L}_t$ is the upper error bound between the ground truth image $\mathbf{x}_0$ and the generated image when we use $\mathbf{x}_0$ as the model input. When we use $\hat{\mathbf{x}}_0$ as the model input and want to get a clear output image, the backward process is almost the same as the backward process when we use $\mathbf{x}_0$ as input because it mainly depends on the neural network parameter. The main difference comes from the forward error. Here, we use $\mathcal{L}_t$ to represent the error between the ground truth image $\mathbf{x}_0$ and the generated image $\tilde{\mathbf{x}}$.

\noindent\textbf{Definition 1.} \textit{\zqh{In the inference period, given the original high-resolution image $\mathbf{x}$ (\ie the ground truth) and the compressed low-resolution version $\hat{\mathbf{x}}$, we first inject $t$ steps of Gaussian noise into $\hat{\mathbf{x}}$ to obtain the noisy version $\hat{\mathbf{x}}_t$, then feed $\hat{\mathbf{x}}_t$ into DGM as the generative seed, and finally, reverse the diffusion process also through $t$ denoising steps to generate the recovered image $\bar{\mathbf{x}}$. 
Therefore, the entire recovery error $\mathcal{L}_t$ between the recovered image $\bar{\mathbf{x}}$ and ground-truth $\mathbf{x}$ can be formulated as:}}
\begin{equation}
    \mathcal{L}_t = D_\text{KL}(q(\mathbf{x}_t \vert \hat{\mathbf{x}}_0) \parallel p_\theta(\mathbf{x}_t)) + \sum_{i=0}^{t-1} L_{i}.\\
\label{Equation:difinationSR}
\end{equation}

\zqh{Based on the definition of the recovery error, our next step is to analyze which are the key terms impacting this error.}

\noindent\textbf{Theorem 1.} \textit{\zqh{The recovery error $\mathcal{L}_t$ can be resolved as two terms: the signature loss $\mathcal{L}^{S}_t$ and fidelity loss $\mathcal{L}^{F}_t$, where the former reflects the structural similarity of the entire visual content while the latter measures the smoothness of detailed textures. The formulation can be described as:}}

\begin{equation}
\begin{aligned}
    \mathcal{L}_t &\triangleq \bar{L}_t + \big[ K_t \parallel \mathbf{x} - \hat{\mathbf{x}}_0 \parallel^2 + A_t \big] \\
    &\triangleq \underbrace{\mathcal{L}^{S}_t}_{\text{\zqh{Signature Loss}}} +  
    \underbrace{ \omega \mathcal{L}^{F}_t}_{\zqh{L_1~\text{Fidelity Loss}}}, \\
\end{aligned}
\label{Equation:SR}
\end{equation}

\zqh{
where $\omega$ is a hyperparameter to guarantee these two loss terms are of the same magnitude. Empirically, we set $\omega=0.004$. Note that $K_t,A_t$ are two intermediate variables decreasing with $t$. Their detailed descriptions and the complete proof can be found in the supplementary material of \S \ref{appendix:A2}. 
This theorem reveals that the signature loss increases with $t$ while the fidelity loss decreases with $t$. A lower signature loss helps preserve visual similarity, \ie in \KMFigure{Figure:compare_celep}, the upscaled image holds a similar human face as the ground truth. Besides, a lower fidelity loss serves as image deblurring, \ie in \KMFigure{Figure:compare_celep}, the upscaled image provides a clear human face with sharp texture. By jointly optimizing these two terms, we can finally upscale the images with high perceptual quality.}

\subsection{Determining Noise Injection via Perceptual Recoverable Field}

\noindent\textbf{Remark.} \textit{To restore the low-resolution image to a high-resolution one with diffusion model, we should inject $t$ steps noise to the low-resolution image $\hat{\mathbf{x}}_0$ so that both $\mathcal{L}_t^S,\omega \mathcal{L}_t^F $ are less than a threshold, then with the capacity of diffusion model, it will restore the noisy blur image $\hat{\mathbf{x}}_t$ to a clear high-quality image $\tilde{\mathbf{x}}$. \zqh{As $t$ controls the amount of injected noise, we call the range of $t$ satisfying the above constraints as  \textbf{Perceptual Recoverable Field (PRF)}. Consequently, searching the PRF corresponds to determining a suitable amount of noise to inject. The searching process can be formulated as solving the following problem.}}

\begin{eqnarray}
\mathop{\arg\min}_{t} \quad \mathcal{L}_t  \nonumber \\
\text{ s.t. } \quad \mathcal{L}_t^S &\leq \mathcal{C}^S \label{Eq:st1}\\
\omega \mathcal{L}_t^F &\leq \mathcal{C}^F, \label{Eq:st2}
\end{eqnarray}

where the first constraint in \KMEquation{Eq:st1} ensures that the injected noise will not destroy the content of the input image and result in a wrong output image. 
Meanwhile, the second constraint in \KMEquation{Eq:st2} ensures that the recovered image $\tilde{\mathbf{x}}$ will become clearer compared with the input low-resolution image $\hat{\mathbf{x}}_0$. $\mathcal{C}^{S}$ and $\mathcal{C}^{F}$ are two constant thresholds.

\begin{figure}
\centering
\includegraphics[width=\linewidth]{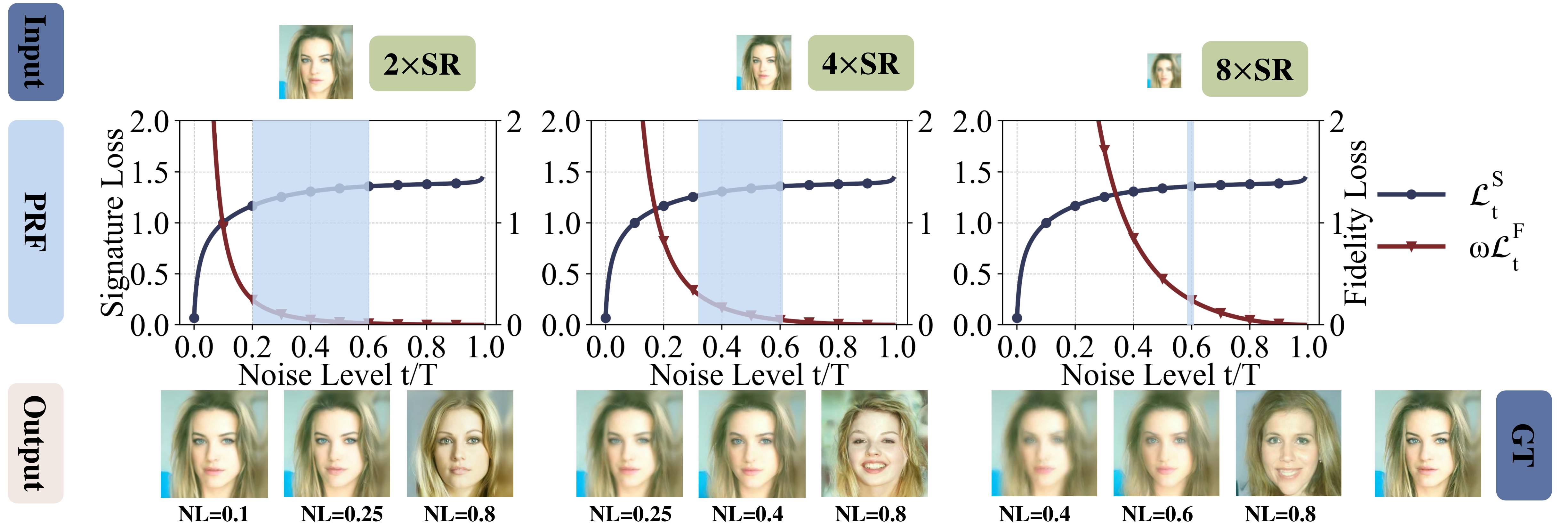}
\caption{The visualization of PRF under different upsampling scales. The PRF satisfies the constraints of both signature and fidelity loss, indicating a suitable amount of noise injection.}
\label{Figure:Error_curve}
\vspace{-15pt}
\end{figure}

We illustrate this decision process in \KMFigure{Figure:Error_curve}, and we use \textbf{NL} here to represent noise level. For different resolutions, the acceptable PRFs are different. For $2\times$ super-resolution, it just needs to inject about $20\%$ noise. Then starting the diffusion sampling process at this point, we can get a high-quality output. For those downsampling in a higher scale like $4\times$ upsampling, we need to inject about $40\%$ noise to get an acceptable output. For $8\times$ upsampling, it doesn't have a PRF area, which means it is hard to restore the original image. Actually, we can still get a reasonable output when we add $50-65\%$ noise to this input. But, compared with other resolutions which have a wider PRF, this output seems to be more blurred and more different in some areas.

\section{Experiments}

\subsection{Experimental Setup}

In our experimental evaluation, we thoroughly investigate the influence of noise injection in different settings. Consistent with the state-of-the-art architecture of DGM, the pre-trained DGM utilizes the U-Net \cite{Unet} architecture as the backbone. For comprehensive information regarding implementation details and hyperparameters, we provide the supplementary material in \S \ref{addition:details}. Our experiments are conducted on four publicly available datasets specifically designed for evaluating image editing tasks: UDM10 \cite{udm10}, REDS \cite{reds}, VID \cite{VID}, and Vimeo90K \cite{vimeo}. Prior to conducting the experiments, we apply preliminary processing steps, including center cropping and resizing the images to a standardized size of $256\times256$. It is important to note that the pre-trained DGM employed in our Diff-SR method is solely trained on images of the $256\times256$ resolution setting and does not have access to any downsampled images. Following the methodology outlined in \cite{SD}, we assess the quality of image editing using both perceptual-based metrics, such as Fréchet Inception Distance (FID) \cite{FID}, which aligns closely with human perception, as well as distortion-based metrics like Peak signal-to-noise ratio (PSNR) \cite{PSNR} and structural similarity index measure (SSIM) \cite{SSIM}. By employing these metrics, we can comprehensively evaluate the impact of noise injection on both fidelity and perception fields \cite{DBLP:journals/csur/LeeVL22}. To establish a fair comparison, we benchmark our method against several baseline methods provided by OpenMMLab \cite{openmmlab}. We train the baseline methods using the same dataset and number of iterations as the pre-trained DGM to ensure fairness in the evaluation. All hyperparameters are set as recommended by OpenMMLab, maintaining experimental consistency.

\subsection{Performance Comparison}

\begin{table*}[htp]
\begin{center}
\caption{Comparison with baseline solutions released by OpenMMLab on UDM10 dataset, where the \redContent{\textbf{red}} and
\blueContent{blue} colors indicate the best and the second-best performance, respectively.} 
\label{Table:inspection_restoration_quality} 
\resizebox{\linewidth}{!}{
\begin{tabular}{*{14}{c}}
\toprule
\multirow{2}*{\makecell[c]{Model}} & \multicolumn{3}{c}{\makecell[c]{$2\times$}} & \multicolumn{3}{c}{\makecell[c]{$2.7\times$}} & \multicolumn{3}{c}{\makecell[c]{$3.5\times$}} & \multicolumn{3}{c}{\makecell[c]{$4\times$}} \\
\cmidrule(lr){2-4}\cmidrule(lr){5-7}\cmidrule(lr){8-10}\cmidrule(lr){11-13}
& FID $\downarrow$ & PSNR $\uparrow$ & SSIM $\uparrow$ & FID $\downarrow$ & PSNR $\uparrow$ & SSIM $\uparrow$ & FID $\downarrow$ & PSNR $\uparrow$ & SSIM $\uparrow$ & FID $\downarrow$ & PSNR $\uparrow$ & SSIM $\uparrow$ \\
\midrule
Nearest \cite{Nearest} & 0.913 & 34.441 & 0.912 & 1.894 & 33.101 & 0.836 & 2.567 & 32.243 & 0.766 & 2.599 & 31.990 & 0.740 \\
Bicubic \cite{Bicubic,Bicubic2} & 1.395 & 35.418 & 0.937 & 3.403 & 33.696 & 0.879 & 5.794 & 32.616 & 0.817 & 6.749 & 32.286 & 0.793 \\
EDSR \cite {EDSR} + Bicubic & \blueContent{\textbf{0.071}} & \redContent{\textbf{40.011}} & \redContent{\textbf{0.979}} & \blueContent{\textbf{0.331}} & \blueContent{\textbf{35.704}} & 0.929 & 0.976 & \blueContent{\textbf{35.308}} & \blueContent{\textbf{0.920}} & 1.470 & 34.445 & 0.893 \\
ESRGAN \cite{ESRGAN} + Bicubic & 0.159 & 38.652 & 0.971 & 0.609 & 35.929 & 0.934 & 1.614 & 34.897 & 0.906 & 1.521 & \blueContent{\textbf{35.073}} & 0.908 \\
LIIF \cite{LIIF} & 0.326 & 38.577 & 0.974 & 1.376 & 35.575 & \blueContent{\textbf{0.941}} & 2.722 & 34.110 & 0.901 & 2.817 & 33.856 & 0.886 \\
SR3 \cite{DBLP:journals/corr/abs-2104-07636} + Bicubic & 1.931 & 27.760 & 0.846 & 0.758 & 29.217 & 0.920 & \blueContent{\textbf{0.464}} & 28.965 & 0.912 & \blueContent{\textbf{0.589}} & 28.770 & \blueContent{\textbf{0.928}} \\
\midrule
\textbf{Diff-SR (ours)} & \redContent{\textbf{0.035}} & \blueContent{\textbf{38.761}} & \blueContent{\textbf{0.975}} & \redContent{\textbf{0.052}} & \redContent{\textbf{37.002}} & \redContent{\textbf{0.963}} & \redContent{\textbf{0.055}} & \redContent{\textbf{37.297}} & \redContent{\textbf{0.967}} & \redContent{\textbf{0.096}} & \redContent{\textbf{35.755}} & \redContent{\textbf{0.957}} \\
\bottomrule
\end{tabular}
}
\vspace{-10pt}
\end{center}
\end{table*}

\begin{table*}[htp]
\begin{center}
\caption{Comparison with baseline solutions released by OpenMMLab on REDS dataset, where the \redContent{\textbf{red}} and
\blueContent{blue} colors indicate the best and the second-best performance, respectively.} 
\label{Table:evaluation_reds}
\resizebox{\linewidth}{!}{
\begin{tabular}{*{14}{c}}
\toprule
\multirow{2}*{\makecell[c]{Model}} & \multicolumn{3}{c}{\makecell[c]{$2\times$}} & \multicolumn{3}{c}{\makecell[c]{$2.7\times$}} & \multicolumn{3}{c}{\makecell[c]{$3.5\times$}} & \multicolumn{3}{c}{\makecell[c]{$4\times$}} \\
\cmidrule(lr){2-4}\cmidrule(lr){5-7}\cmidrule(lr){8-10}\cmidrule(lr){11-13}
& FID $\downarrow$ & PSNR $\uparrow$ & SSIM $\uparrow$ & FID $\downarrow$ & PSNR $\uparrow$ & SSIM $\uparrow$ & FID $\downarrow$ & PSNR $\uparrow$ & SSIM $\uparrow$ & FID $\downarrow$ & PSNR $\uparrow$ & SSIM $\uparrow$ \\
\midrule
Nearest \cite{Nearest} & 2.810 & 32.659 & 0.847 & 3.021 & 31.870 & 0.764 & 3.985 & 31.229 & 0.680 & 4.410 & 31.003 & 0.645 \\
Bicubic \cite{Bicubic,Bicubic2} & 4.756 & 32.833 & 0.870 & 7.375 & 32.074 & 0.808 & 11.289 & 31.362 & 0.729 & 13.192 & 31.090 & 0.694 \\
EDSR \cite{EDSR} + Bicubic & \redContent{\textbf{0.182}} & 32.678 & 0.875 & 1.824 & 30.827 & 0.719 & 1.748 & 30.298 & 0.627 & 3.172 & 29.992 & 0.570 \\
ESRGAN \cite{ESRGAN} + Bicubic & 1.422 & 33.011 & 0.893 & 32.309 & 27.843 & 0.119 & 28.127 & 27.843 & 0.113 & 24.467 & 27.881 & 0.113 \\
LIIF \cite{LIIF} & 2.204 & \redContent{\textbf{34.114}} & \redContent{\textbf{0.926}} & 4.481 & \blueContent{\textbf{32.822}} & \blueContent{\textbf{0.863}} & 7.946 & \blueContent{\textbf{31.935}} & \blueContent{\textbf{0.796}} & 7.754 & \blueContent{\textbf{31.867}} & \blueContent{\textbf{0.791}} \\
SR3 \cite{DBLP:journals/corr/abs-2104-07636} + Bicubic & \blueContent{\textbf{0.421}} & 29.356 & 0.891 & \blueContent{\textbf{0.939}} & 28.355 & 0.529 & \blueContent{\textbf{1.053}} & 28.329 & 0.506 & \redContent{\textbf{0.683}} & 28.699 & 0.765 \\
\midrule
\textbf{Diff-SR} & 0.810 & \blueContent{\textbf{33.671}} & \blueContent{\textbf{0.911}} & \redContent{\textbf{0.388}} & \redContent{\textbf{33.188}} & \redContent{\textbf{0.889}} & \redContent{\textbf{0.432}} & \redContent{\textbf{32.972}} & \redContent{\textbf{0.882}} & \blueContent{\textbf{0.918}} & \redContent{\textbf{32.441}} & \redContent{\textbf{0.879}} \\
\bottomrule
\end{tabular}
} 
\vspace{-15pt}
\end{center}
\end{table*}

\begin{figure}[htp]
  \centering
    \includegraphics[width=1\linewidth,height=2.15in]{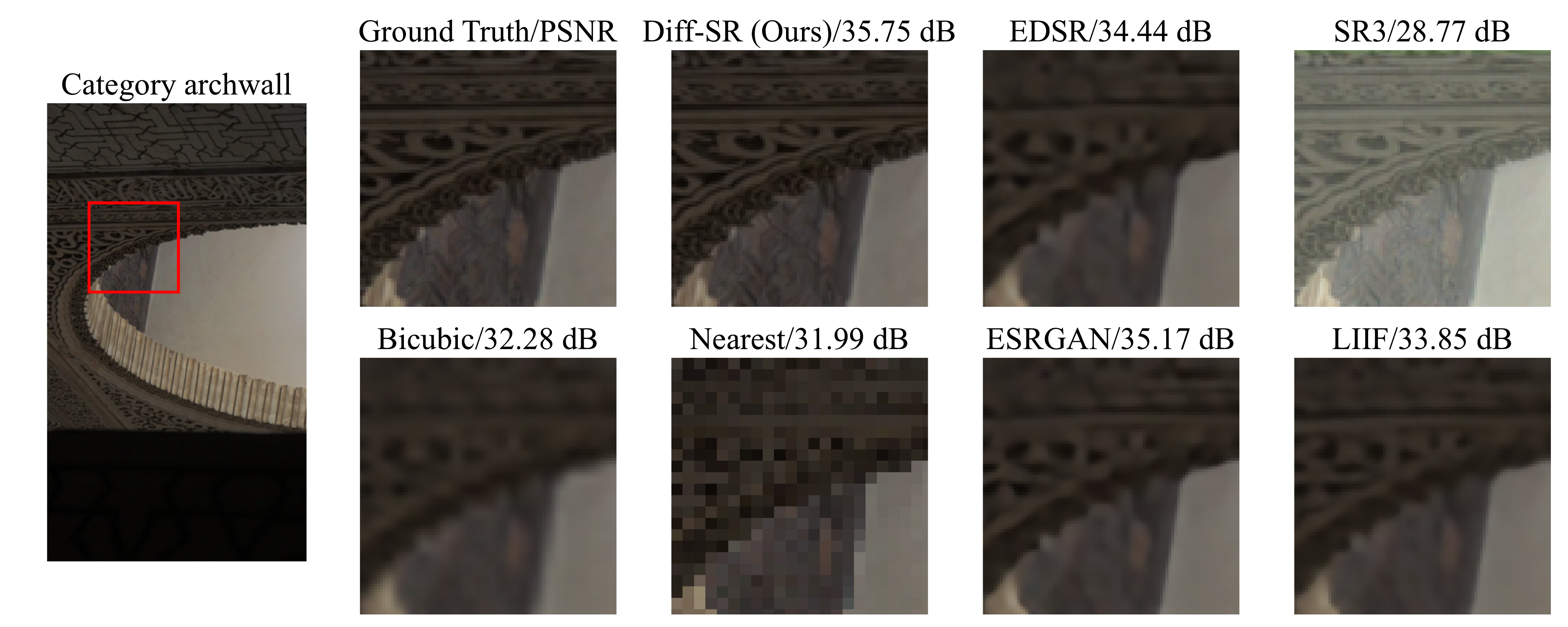}
  \caption{Perfromance visualization of different $4\times$ SR solutions, evaluated on UDM10 dataset. Additional results are provided in the supplementary material of \S \ref{section:addition_images}. (\textbf{Zoom-in for best view})}
  \label{Figure:illustrate_images}
\vspace{-5pt}
\end{figure}

As demonstrated in \KMTable{Table:inspection_restoration_quality} and \KMTable{Table:evaluation_reds} (with additional results provided in the supplementary material of Section \ref{Section:other_tables}), our model consistently achieves superior performance across the majority of datasets. At lower compression rates (e.g., 2$\times$), while Diff-SR delivers commendable overall metrics, its advantage over certain baselines may not be substantial, as some alternatives manage to achieve similar performance levels. However, as the compression rate increases (e.g., 4$\times$), the distinct superiority of Diff-SR becomes more evident. Higher compression rates inevitably result in the loss or degradation of original information, leading to visual artifacts, blurring, and other forms of distortion. Consequently, restoring images for other baselines becomes more challenging, even after retraining these models on the specific data. In contrast, Diff-SR maintains almost identical performance, even under high-rate compression scenarios. Notably, Diff-SR exhibits more significant improvements in perception field metrics, such as FID, compared to fidelity field metrics like PSNR and SSIM. This discrepancy arises from the fact that PSNR and SSIM tend to favor mean squared error (MSE) regression-based techniques, which tend to be excessively conservative with high-frequency details. However, these metrics penalize synthetic high-frequency details that may not align well with the target image \cite{DBLP:journals/corr/abs-2104-07636}. Consequently, diffusion-based models like SR3 and Diff-SR achieve higher FID scores while occasionally exhibiting lower PSNR and SSIM scores. The performance advantage of Diff-SR is further verified through visualizations, as illustrated in \KMFigure{Figure:illustrate_images}.

Importantly, when considering other baselines, achieving upsampling scales such as 2.7$\times$ poses certain challenges. One approach involves employing a 2$\times$ super-resolution model to enhance an image from $96\times96$ to $192\times192$ and subsequently using basic upsampling methods like Bicubic to upscale it to $256\times256$. Alternatively, another method involves using a 4$\times$ super-resolution model to enhance the image and then downsampling the output to $256\times256$. However, both of these methods suffer from performance limitations due to the distortions introduced by further compression and the mismatch of compression scales. In contrast, our Diff-SR approach leverages a single pre-trained model to handle diverse scale super-resolution tasks. By simply adjusting the noise injection level, Diff-SR achieves excellent results for various scale SR scenarios. Notably, Diff-SR is capable of directly accepting image inputs of any scale (e.g., 2.7$\times$, 3.5$\times$) without relying on techniques like Bicubic interpolation to conduct post-process. Furthermore, it is worth mentioning that due to the inherent instability of adversarial training, ESRGAN may encounter issues such as model collapse \cite{GANcrack, GANcrack2} on certain datasets, leading to the failure to reconstruct LR images. This highlights the advantages of Diff-SR over ESRGAN in terms of stability and reliable image reconstruction.

\subsection{Ablation Studies}
\subsubsection{Impact of Upsampling Scales}

\begin{figure}[htp]
\vspace{-5pt}
  \centering
    \includegraphics[width=\linewidth,height=1.37in]{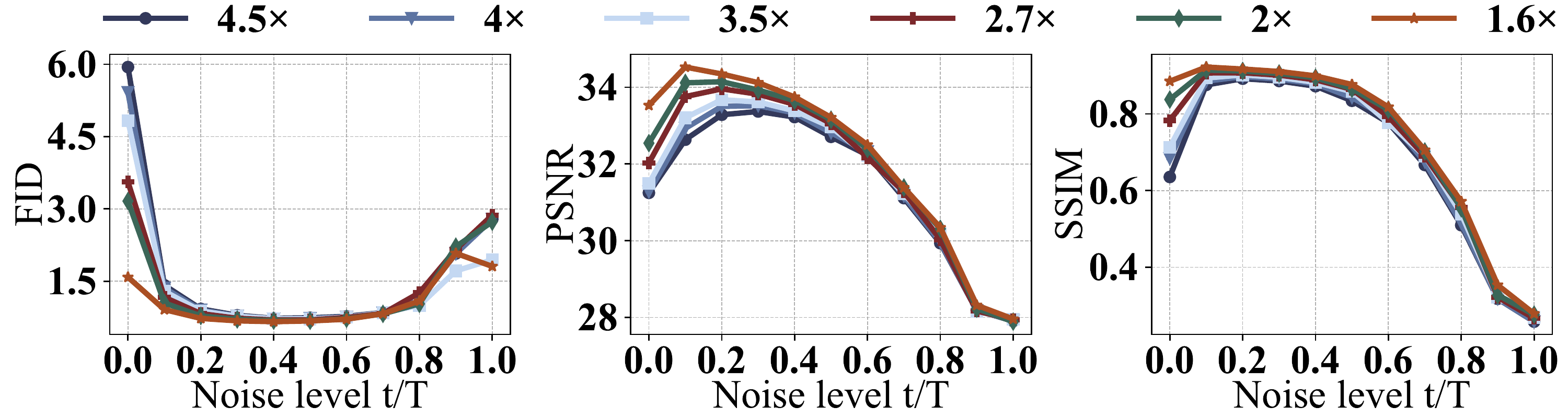}
  \caption{Ablation study of upsampling scales on REDS dataset.}
  \label{Figure:experiment_noise_level}
\vspace{-10pt}
\end{figure}

In the conducted ablation experiment, as illustrated in \KMFigure{Figure:experiment_noise_level}, we explored various super-resolution (SR) scales ranging from 1.6$\times$ to 4.5$\times$. The results revealed the necessity of adapting noise injection levels to suit different SR tasks. Initially, at the state $t/T=0$, we observed a deterioration in metrics such as FID, PSNR, and SSIM as the resolution decreased. However, by employing Diff-SR to restore these images, their quality experienced significant improvement, particularly when the noise level was not excessively high. \KMFigure{Figure:experiment_noise_level} presents the outcomes of the experiment. For a 2$\times$ upscaling SR task, Diff-SR achieved FID (0.50), PSNR (34.2 dB), and SSIM (89.5) scores by employing a noise injection level of approximately 20\%. Conversely, to attain similar performance in a 4$\times$ upscaling SR task, Diff-SR required a noise injection level of approximately 40\%, resulting in FID (0.55), PSNR (33.2 dB), and SSIM (88.4) scores. The observed discrepancies in these metrics are primarily attributable to the degradation of the input image. Nevertheless, when compared to the initial state, Diff-SR substantially mitigated these discrepancies. Importantly, this experiment emphasized the importance of avoiding excessive noise injection into the input images. As depicted in the figure, if the noise injection level exceeds 60\%, all metrics exhibit a deterioration worse than that of the initial state. This finding aligns with our preliminary observations, where excessively high noise injection levels led to visually clear output images but compromised their semantic content.

\subsubsection{Impact of DGM Pre-training Scheduler}

\begin{figure}[htp]
\vspace{-5pt}
  \centering
    \includegraphics[width=\linewidth,height=1.37in]{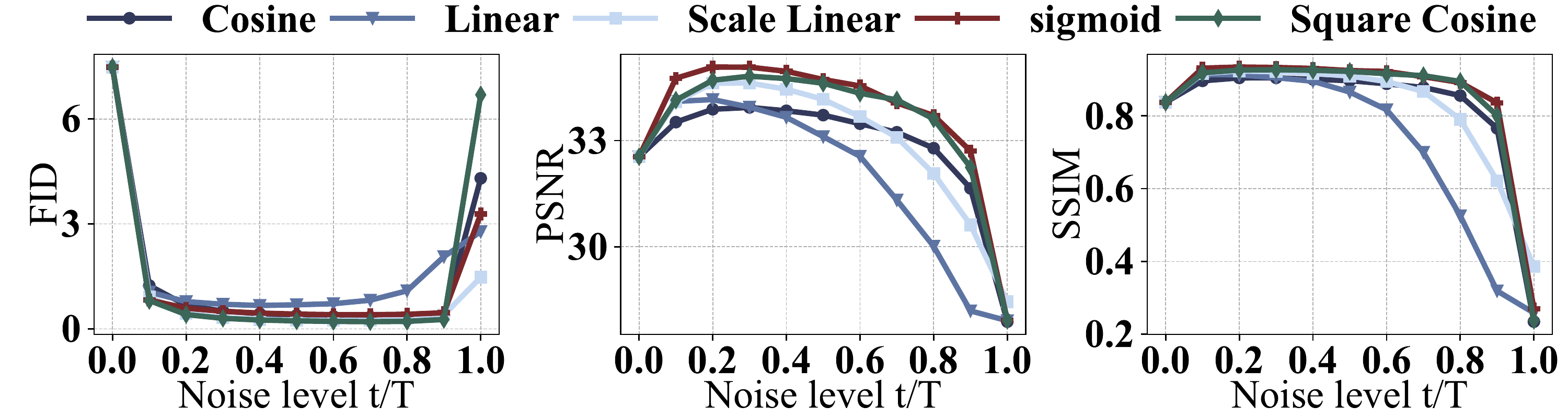}
  \caption{Ablation study of DGM pre-training scheduler on REDS dataset.}
  \label{Figure:ablation_schedular}
\vspace{-20pt}
\end{figure}

We further conducted an ablation study to investigate the impact of different hyperparameters on the required noise injection levels for achieving target-scale super-resolution. Notably, we identified the noise scheduler method as a significant influencing factor. The noise scheduler method regulates the strength of noise injection at each step within the DGM model, which is accomplished by manipulating the parameters $\alpha_t$ and $\beta_t$. To explore this, we examined several noise schedulers, including Cosine \cite{cosine}, Linear \cite{linear}, Scale Linear \cite{scalelinear}, Sigmoid \cite{sigmoid}, and Square Cosine \cite{SD}, which have been previously adopted in Stable Diffusion \cite{SD}. For this ablation experiment, we focused on a 4$\times$ super-resolution scale. The results of this study, presented in \KMFigure{Figure:ablation_schedular}, showcased the behavior of both methods starting from the same initial metric point and exhibited similar trends as the noise injection level varied. Among the various scheduler methods, the Square Cosine scheduler emerged as the most effective. It achieved a superior FID score (0.2), the second-best PSNR score (34.2 dB), and SSIM score (89\%) within a noise injection level range of 10\% to 70\%. In contrast, schedulers such as Cosine and Linear, which were initially employed in the original DGM version \cite{DBLP:conf/nips/HoJA20}, demonstrated poorer performance on these metrics at the same noise injection level. Additionally, the acceptable range of noise levels for satisfactory results was narrower compared to the Square Cosine scheduler.

\section{Conclusion}

This research endeavor aims to explore the arbitrary-scale super-resolution (ASSR) capabilities inherent in existing pre-trained diffusion-based generative models (DGMs), without necessitating additional fine-tuning or distillation efforts. We present Diff-SR, the pioneering ASSR approach that relies solely on pre-trained DGMs. The foundation of Diff-SR lies in a simple yet powerful observation: by introducing a specific level of noise into the low-resolution (LR) image prior to initiating the DGM's backward diffusion process, the desired recovery performance can be achieved. We further substantiate the feasibility of this methodology through theoretical analysis and introduce a fundamental metric known as the \textit{Perceptual Recoverable Field} (PRF). The PRF metric quantifies the permissible amount of noise that can be injected to ensure high-quality recovery for various upsampling scales. Our extensive experiments demonstrate that Diff-SR surpasses existing state-of-the-art solutions across diverse super-resolution scenarios.

\clearpage

{
\small
\bibliographystyle{plain}
\bibliography{diffsr}

\begin{thebibliography}{10}

\bibitem{Bicubic}
Francesc Ar{\`{a}}ndiga.
\newblock A nonlinear algorithm for monotone piecewise bicubic interpolation.
\newblock {\em Appl. Math. Comput.}, 272:100--113, 2016.

\bibitem{scalelinear}
Jacob Austin, Daniel~D. Johnson, Jonathan Ho, Daniel Tarlow, and Rianne van~den
  Berg.
\newblock Structured denoising diffusion models in discrete state-spaces.
\newblock In Marc'Aurelio Ranzato, Alina Beygelzimer, Yann~N. Dauphin, Percy
  Liang, and Jennifer~Wortman Vaughan, editors, {\em Advances in Neural
  Information Processing Systems 34: Annual Conference on Neural Information
  Processing Systems 2021, NeurIPS 2021, December 6-14, 2021, virtual}, pages
  17981--17993, 2021.

\bibitem{GANcrack}
David Bau, Jun{-}Yan Zhu, Jonas Wulff, William~S. Peebles, Bolei Zhou, Hendrik
  Strobelt, and Antonio Torralba.
\newblock Seeing what a {GAN} cannot generate.
\newblock In {\em 2019 {IEEE/CVF} International Conference on Computer Vision,
  {ICCV} 2019, Seoul, Korea (South), October 27 - November 2, 2019}, pages
  4501--4510. {IEEE}, 2019.

\bibitem{Fourier}
Ronald~Newbold Bracewell and Ronald~N Bracewell.
\newblock {\em The Fourier transform and its applications}, volume 31999.
\newblock McGraw-Hill New York, 1986.

\bibitem{Fourier2}
E~Oran Brigham and RE~Morrow.
\newblock The fast fourier transform.
\newblock {\em IEEE spectrum}, 4(12):63--70, 1967.

\bibitem{sigmoid}
Ting Chen.
\newblock On the importance of noise scheduling for diffusion models.
\newblock {\em CoRR}, abs/2301.10972, 2023.

\bibitem{LIIF}
Yinbo Chen, Sifei Liu, and Xiaolong Wang.
\newblock Learning continuous image representation with local implicit image
  function.
\newblock In {\em Proceedings of the IEEE/CVF Conference on Computer Vision and
  Pattern Recognition}, pages 8628--8638, 2021.

\bibitem{Markovchain}
Wai-Ki Ching and Michael~K Ng.
\newblock Markov chains.
\newblock {\em Models, algorithms and applications}, 2006.

\bibitem{VideoDFM3}
Florinel{-}Alin Croitoru, Vlad Hondru, Radu~Tudor Ionescu, and Mubarak Shah.
\newblock Diffusion models in vision: {A} survey.
\newblock {\em CoRR}, abs/2209.04747, 2022.

\bibitem{DbeatGAN}
Prafulla Dhariwal and Alexander~Quinn Nichol.
\newblock Diffusion models beat gans on image synthesis.
\newblock In Marc'Aurelio Ranzato, Alina Beygelzimer, Yann~N. Dauphin, Percy
  Liang, and Jennifer~Wortman Vaughan, editors, {\em Advances in Neural
  Information Processing Systems 34: Annual Conference on Neural Information
  Processing Systems 2021, NeurIPS 2021, December 6-14, 2021, virtual}, pages
  8780--8794, 2021.

\bibitem{DBLP:conf/nips/DhariwalN21}
Prafulla Dhariwal and Alexander~Quinn Nichol.
\newblock Diffusion models beat gans on image synthesis.
\newblock In Marc'Aurelio Ranzato, Alina Beygelzimer, Yann~N. Dauphin, Percy
  Liang, and Jennifer~Wortman Vaughan, editors, {\em Proceedings of the Annual
  Conference on Neural Information Processing Systems (NeurIPS)}, pages
  8780--8794, 2021.

\bibitem{finetuning2}
Rinon Gal, Yuval Alaluf, Yuval Atzmon, Or~Patashnik, Amit~H Bermano, Gal
  Chechik, and Daniel Cohen-Or.
\newblock An image is worth one word: Personalizing text-to-image generation
  using textual inversion.
\newblock {\em arXiv preprint arXiv:2208.01618}, 2022.

\bibitem{SR2}
Muhammad Haris, Greg Shakhnarovich, and Norimichi Ukita.
\newblock Erratum to "deep back-projection networks for single image
  super-resolution".
\newblock {\em {IEEE} Trans. Pattern Anal. Mach. Intell.}, 44(2):1122, 2022.

\bibitem{DFMlongV}
William Harvey, Saeid Naderiparizi, Vaden Masrani, Christian Weilbach, and
  Frank Wood.
\newblock Flexible diffusion modeling of long videos.
\newblock In {\em NeurIPS}, 2022.

\bibitem{KL2}
John~R. Hershey and Peder~A. Olsen.
\newblock Approximating the kullback leibler divergence between gaussian
  mixture models.
\newblock In {\em Proceedings of the {IEEE} International Conference on
  Acoustics, Speech, and Signal Processing, {ICASSP} 2007, Honolulu, Hawaii,
  USA, April 15-20, 2007}, pages 317--320. {IEEE}, 2007.

\bibitem{FID}
Martin Heusel, Hubert Ramsauer, Thomas Unterthiner, Bernhard Nessler, and Sepp
  Hochreiter.
\newblock Gans trained by a two time-scale update rule converge to a local nash
  equilibrium.
\newblock In Isabelle Guyon, Ulrike von Luxburg, Samy Bengio, Hanna~M. Wallach,
  Rob Fergus, S.~V.~N. Vishwanathan, and Roman Garnett, editors, {\em Advances
  in Neural Information Processing Systems 30: Annual Conference on Neural
  Information Processing Systems 2017, December 4-9, 2017, Long Beach, CA,
  {USA}}, pages 6626--6637, 2017.

\bibitem{DBLP:conf/nips/HoJA20}
Jonathan Ho, Ajay Jain, and Pieter Abbeel.
\newblock Denoising diffusion probabilistic models.
\newblock In {\em Proceedings of the Annual Conference on Neural Information
  Processing Systems (NeurIPS)}, 2020.

\bibitem{PSNR}
Alain Hor{\'{e}} and Djemel Ziou.
\newblock Image quality metrics: {PSNR} vs. {SSIM}.
\newblock In {\em 20th International Conference on Pattern Recognition, {ICPR}
  2010, Istanbul, Turkey, 23-26 August 2010}, pages 2366--2369. {IEEE} Computer
  Society, 2010.

\bibitem{finetuning}
Edward~J Hu, Yelong Shen, Phillip Wallis, Zeyuan Allen-Zhu, Yuanzhi Li, Shean
  Wang, Lu~Wang, and Weizhu Chen.
\newblock Lora: Low-rank adaptation of large language models.
\newblock {\em arXiv preprint arXiv:2106.09685}, 2021.

\bibitem{ASSR}
Zheng Hui, Xinbo Gao, Yunchu Yang, and Xiumei Wang.
\newblock Lightweight image super-resolution with information
  multi-distillation network.
\newblock In Laurent Amsaleg, Benoit Huet, Martha~A. Larson, Guillaume Gravier,
  Hayley Hung, Chong{-}Wah Ngo, and Wei~Tsang Ooi, editors, {\em Proceedings of
  the 27th {ACM} International Conference on Multimedia, {MM} 2019, Nice,
  France, October 21-25, 2019}, pages 2024--2032. {ACM}, 2019.

\bibitem{Bicubic2}
Jung~Woo Hwang and Hwang~Soo Lee.
\newblock Adaptive image interpolation based on local gradient features.
\newblock {\em {IEEE} Signal Process. Lett.}, 11(3):359--362, 2004.

\bibitem{KLdiver}
James~M Joyce.
\newblock Kullback-leibler divergence.
\newblock In {\em International encyclopedia of statistical science}, pages
  720--722. Springer, 2011.

\bibitem{DDGM}
Bahjat Kawar, Michael Elad, Stefano Ermon, and Jiaming Song.
\newblock Denoising diffusion restoration models.
\newblock In {\em NeurIPS}, 2022.

\bibitem{Adam}
Diederik~P. Kingma and Jimmy Ba.
\newblock Adam: {A} method for stochastic optimization.
\newblock In {\em Proceedings of the International Conference on Learning
  Representations (ICLR)}, 2015.

\bibitem{linear}
Diederik~P. Kingma, Tim Salimans, Ben Poole, and Jonathan Ho.
\newblock Variational diffusion models.
\newblock {\em CoRR}, abs/2107.00630, 2021.

\bibitem{SR1}
Christian Ledig, Lucas Theis, Ferenc Huszar, Jose Caballero, Andrew Cunningham,
  Alejandro Acosta, Andrew~P. Aitken, Alykhan Tejani, Johannes Totz, Zehan
  Wang, and Wenzhe Shi.
\newblock Photo-realistic single image super-resolution using a generative
  adversarial network.
\newblock In {\em 2017 {IEEE} Conference on Computer Vision and Pattern
  Recognition, {CVPR} 2017, Honolulu, HI, USA, July 21-26, 2017}, pages
  105--114. {IEEE} Computer Society, 2017.

\bibitem{DBLP:journals/csur/LeeVL22}
Royson Lee, Stylianos~I. Venieris, and Nicholas~D. Lane.
\newblock Deep neural network-based enhancement for image and video streaming
  systems: {A} survey and future directions.
\newblock {\em {ACM} Comput. Surv.}, 54(8):169:1--169:30, 2022.

\bibitem{EDSR}
Bee Lim, Sanghyun Son, Heewon Kim, Seungjun Nah, and Kyoung Mu~Lee.
\newblock Enhanced deep residual networks for single image super-resolution.
\newblock In {\em Proceedings of the IEEE conference on computer vision and
  pattern recognition workshops}, pages 136--144, 2017.

\bibitem{Nearest}
Chung{-}Chi Lin, Ming{-}Hwa Sheu, Huann{-}Keng Chiang, Chishyan Liaw, and
  Zeng{-}Chuan Wu.
\newblock The efficient {VLSI} design of {BI-CUBIC} convolution interpolation
  for digital image processing.
\newblock In {\em International Symposium on Circuits and Systems {(ISCAS}
  2008), 18-21 May 2008, Sheraton Seattle Hotel, Seattle, Washington, {USA}},
  pages 480--483. {IEEE}, 2008.

\bibitem{VID}
Ce~Liu and Deqing Sun.
\newblock On bayesian adaptive video super resolution.
\newblock {\em IEEE Transactions on Pattern Analysis and Machine Intelligence},
  36(2):346--360, 2013.

\bibitem{Distillation3}
Eric Luhman and Troy Luhman.
\newblock Knowledge distillation in iterative generative models for improved
  sampling speed.
\newblock {\em CoRR}, abs/2101.02388, 2021.

\bibitem{reds}
Seungjun Nah, Sungyong Baik, Seokil Hong, Gyeongsik Moon, Sanghyun Son, Radu
  Timofte, and Kyoung~Mu Lee.
\newblock Ntire 2019 challenge on video deblurring and super-resolution:
  Dataset and study.
\newblock In {\em CVPR Workshops}, June 2019.

\bibitem{cosine}
Alexander~Quinn Nichol and Prafulla Dhariwal.
\newblock Improved denoising diffusion probabilistic models.
\newblock In Marina Meila and Tong Zhang, editors, {\em Proceedings of the 38th
  International Conference on Machine Learning, {ICML} 2021, 18-24 July 2021,
  Virtual Event}, volume 139 of {\em Proceedings of Machine Learning Research},
  pages 8162--8171. {PMLR}, 2021.

\bibitem{openmmlab}
OpenMMLab.
\newblock Openmmlab dataset.
\newblock \url{https://openmmlab.com/dataset}, 2022.

\bibitem{DallE}
Aditya Ramesh, Prafulla Dhariwal, Alex Nichol, Casey Chu, and Mark Chen.
\newblock Hierarchical text-conditional image generation with {CLIP} latents.
\newblock {\em CoRR}, abs/2204.06125, 2022.

\bibitem{finetuning3}
Daniel Roich, Ron Mokady, Amit~H Bermano, and Daniel Cohen-Or.
\newblock Pivotal tuning for latent-based editing of real images.
\newblock {\em ACM Transactions on Graphics (TOG)}, 42(1):1--13, 2022.

\bibitem{SD}
Robin Rombach, Andreas Blattmann, Dominik Lorenz, Patrick Esser, and
  Bj{\"{o}}rn Ommer.
\newblock High-resolution image synthesis with latent diffusion models.
\newblock In {\em {IEEE/CVF} Conference on Computer Vision and Pattern
  Recognition, {CVPR} 2022, New Orleans, LA, USA, June 18-24, 2022}, pages
  10674--10685. {IEEE}, 2022.

\bibitem{Unet}
Olaf Ronneberger.
\newblock Invited talk: U-net convolutional networks for biomedical image
  segmentation.
\newblock In Klaus~Hermann Maier{-}Hein, Thomas~Martin Deserno, Heinz Handels,
  and Thomas Tolxdorff, editors, {\em Bildverarbeitung f{\"{u}}r die Medizin
  2017 - Algorithmen - Systeme - Anwendungen. Proceedings des Workshops vom 12.
  bis 14. M{\"{a}}rz 2017 in Heidelberg}, Informatik Aktuell, page~3. Springer,
  2017.

\bibitem{DBLP:journals/corr/abs-2104-07636}
Chitwan Saharia, Jonathan Ho, William Chan, Tim Salimans, David~J. Fleet, and
  Mohammad Norouzi.
\newblock Image super-resolution via iterative refinement.
\newblock {\em CoRR}, abs/2104.07636, 2021.

\bibitem{Distillation1}
Tim Salimans and Jonathan Ho.
\newblock Progressive distillation for fast sampling of diffusion models.
\newblock In {\em The Tenth International Conference on Learning
  Representations, {ICLR} 2022, Virtual Event, April 25-29, 2022}.
  OpenReview.net, 2022.

\bibitem{DBLP:conf/icml/Sohl-DicksteinW15}
Jascha Sohl{-}Dickstein, Eric~A. Weiss, Niru Maheswaranathan, and Surya
  Ganguli.
\newblock Deep unsupervised learning using nonequilibrium thermodynamics.
\newblock In Francis~R. Bach and David~M. Blei, editors, {\em Proceedings of
  the 32nd International Conference on Machine Learning, {ICML} 2015, Lille,
  France, 6-11 July 2015}, volume~37 of {\em {JMLR} Workshop and Conference
  Proceedings}, pages 2256--2265. JMLR.org, 2015.

\bibitem{DBLP:conf/iclr/SongME21}
Jiaming Song, Chenlin Meng, and Stefano Ermon.
\newblock Denoising diffusion implicit models.
\newblock In {\em 9th International Conference on Learning Representations,
  {ICLR} 2021, Virtual Event, Austria, May 3-7, 2021}. OpenReview.net, 2021.

\bibitem{Distillation2}
Yang Song, Prafulla Dhariwal, Mark Chen, and Ilya Sutskever.
\newblock Consistency models.
\newblock {\em CoRR}, abs/2303.01469, 2023.

\bibitem{SDE}
Yang Song, Jascha Sohl{-}Dickstein, Diederik~P. Kingma, Abhishek Kumar, Stefano
  Ermon, and Ben Poole.
\newblock Score-based generative modeling through stochastic differential
  equations.
\newblock In {\em 9th International Conference on Learning Representations,
  {ICLR} 2021, Virtual Event, Austria, May 3-7, 2021}. OpenReview.net, 2021.

\bibitem{LIIF2}
Longguang Wang, Yingqian Wang, Zaiping Lin, Jungang Yang, Wei An, and Yulan
  Guo.
\newblock Learning {A} single network for scale-arbitrary super-resolution.
\newblock In {\em 2021 {IEEE/CVF} International Conference on Computer Vision,
  {ICCV} 2021, Montreal, QC, Canada, October 10-17, 2021}, pages 4781--4790.
  {IEEE}, 2021.

\bibitem{ESRGAN2}
Xintao Wang, Liangbin Xie, Chao Dong, and Ying Shan.
\newblock Real-esrgan: Training real-world blind super-resolution with pure
  synthetic data.
\newblock {\em CoRR}, abs/2107.10833, 2021.

\bibitem{ESRGAN}
Xintao Wang, Ke~Yu, Shixiang Wu, Jinjin Gu, Yihao Liu, Chao Dong, Yu~Qiao, and
  Chen~Change Loy.
\newblock {ESRGAN:} enhanced super-resolution generative adversarial networks.
\newblock In Laura Leal{-}Taix{\'{e}} and Stefan Roth, editors, {\em Computer
  Vision - {ECCV} 2018 Workshops - Munich, Germany, September 8-14, 2018,
  Proceedings, Part {V}}, volume 11133 of {\em Lecture Notes in Computer
  Science}, pages 63--79. Springer, 2018.

\bibitem{SSIM}
Zhou Wang, Alan~C. Bovik, Hamid~R. Sheikh, and Eero~P. Simoncelli.
\newblock Image quality assessment: from error visibility to structural
  similarity.
\newblock {\em {IEEE} Trans. Image Process.}, 13(4):600--612, 2004.

\bibitem{weng2021diffusion}
Lilian Weng.
\newblock What are diffusion models?
\newblock {\em lilianweng.github.io}, Jul 2021.

\bibitem{LIIF3}
Xingqian Xu, Zhangyang Wang, and Humphrey Shi.
\newblock Ultrasr: Spatial encoding is a missing key for implicit image
  function-based arbitrary-scale super-resolution.
\newblock {\em CoRR}, abs/2103.12716, 2021.

\bibitem{vimeo}
Tianfan Xue, Baian Chen, Jiajun Wu, Donglai Wei, and William~T Freeman.
\newblock Video enhancement with task-oriented flow.
\newblock {\em International Journal of Computer Vision (IJCV)},
  127(8):1106--1125, 2019.

\bibitem{SISR2}
Chih{-}Yuan Yang, Chao Ma, and Ming{-}Hsuan Yang.
\newblock Single-image super-resolution: {A} benchmark.
\newblock In David~J. Fleet, Tom{\'{a}}s Pajdla, Bernt Schiele, and Tinne
  Tuytelaars, editors, {\em Computer Vision - {ECCV} 2014 - 13th European
  Conference, Zurich, Switzerland, September 6-12, 2014, Proceedings, Part
  {IV}}, volume 8692 of {\em Lecture Notes in Computer Science}, pages
  372--386. Springer, 2014.

\bibitem{SISR}
Wenming Yang, Xuechen Zhang, Yapeng Tian, Wei Wang, and Jing{-}Hao Xue.
\newblock Deep learning for single image super-resolution: {A} brief review.
\newblock {\em CoRR}, abs/1808.03344, 2018.

\bibitem{udm10}
Peng Yi, Zhongyuan Wang, Kui Jiang, Junjun Jiang, and Jiayi Ma.
\newblock Progressive fusion video super-resolution network via exploiting
  non-local spatio-temporal correlations.
\newblock In {\em IEEE International Conference on Computer Vision (ICCV)},
  pages 3106--3115, 2019.

\bibitem{EDSR2}
Jiahui Yu, Yuchen Fan, Jianchao Yang, Ning Xu, Zhaowen Wang, Xinchao Wang, and
  Thomas~S. Huang.
\newblock Wide activation for efficient and accurate image super-resolution.
\newblock {\em CoRR}, abs/1808.08718, 2018.

\bibitem{GANcrack2}
Zhaoyu Zhang, Mengyan Li, and Jun Yu.
\newblock On the convergence and mode collapse of {GAN}.
\newblock In Nafees~Bin Zafar and Kun Zhou, editors, {\em {SIGGRAPH} Asia 2018
  Technical Briefs, Tokyo, Japan, December 04-07, 2018}, pages 21:1--21:4.
  {ACM}, 2018.

\end{thebibliography}
}

\appendix

\section{Notations}

\begin{table}[htbp]
\centering
\caption{Notation list.}
\label{Table:Notation List}
\begin{tabular}{ll}
    \toprule
    Notation& Description\\
    \midrule
    $\beta_t$ & The variance schedule of Diffusion model $\beta_t \in (0,1),\text{where}~t \in [1,T]$ \\
    $\alpha_t$ & Defined as $\alpha_t = 1 - \beta_t$ \\
    $\bar{\alpha}_t$ & Defined as $\bar{\alpha}_t = \prod_{i=1}^t \alpha_i$ \\
    $\tilde{\beta}_t$ & The output based on $\bar{\alpha_t},\beta_t$, $\tilde{\beta}_t = \frac{1 - \bar{\alpha}_{t-1}}{1 - \bar{\alpha}_t} \cdot \beta_t$ \\
    $\mathbf{x}$ & The original high-resolution image, \ie the ground truth \\
    $\mathbf{x}_0$ & The initial status before injecting Gaussian noise, \ie $\mathbf{x}_0=\mathbf{x}$ \\
    $\mathbf{x}_t$ & The final status after injecting Gaussian noise by $t$ steps to ground truth \\
    $\hat{\mathbf{x}}$ & The low-resolution image compressed from $\mathbf{x}$ \\
    $\bar{\mathbf{x}}$ & The recovered image based on $\hat{\mathbf{x}}$ \\
    $t$ & The step number that injects Gaussian noise into $\hat{x}$, $t \in [0,T]$ \\
    $\hat{\mathbf{x}}_0$ & The initial status before injecting Gaussian noise, \ie $\hat{\mathbf{x}}_0=\hat{\mathbf{x}}$ \\
    $\hat{\mathbf{x}}_t$ & The final status after injecting Gaussian noise by $t$ steps \\
    $\tilde{\mathbf{\mu}_t}$ & The mean of the reverse gaussian distribution \\
    $\epsilon_t$ & The injected noise at timestep $t$ \\
    $\epsilon_{\theta}$ & The predicted noise by neural network \\
    $E_t$ & The prediction error between $\epsilon_t$ and $\epsilon_{\theta}$ \\
    $\bar{L}_t$ & The accumulation error of $t$ steps denoising with $\mathbf{x}$ as input \\
    $\mathcal{L}_t$ & The accumulation error of $t$ steps denoising with $\hat{\mathbf{x}}$ as input \\
    $A_t,K_t$ & Derived intermediate variables decrease along with $t$ \\
    $\mathcal{L}^{S}_t$ & The signature loss \\
    $\mathcal{L}^{F}_t$ & The fidelity loss \\
    $\omega$ & The hyperparameter that guarantees $\mathcal{L}^{S}_t$ and $\mathcal{L}^{F}_t$ with the same order of magnitude \\
    
    \bottomrule
\end{tabular}
\end{table}

All the notations used in the supplementary material are listed in \KMTable{Table:Notation List}.

\section{Proof of \textbf{Lemma 1.} in Sec. 3.2}
\label{appendix:lemma1}

\begin{proof}

Following the idea of DDPM, when we start the reverse process at time step $t$ the variational upper bound to optimize the negative log-likelihood can be rewritten as:
\begin{equation}
\begin{aligned}
- \log p_\theta(\mathbf{x}_0) 
&\leq - \log p_\theta(\mathbf{x}_0) + D_\text{KL}(q(\mathbf{x}_{1:t}\vert\mathbf{x}_0) \| p_\theta(\mathbf{x}_{1:t}\vert\mathbf{x}_0) ) \\
&= \mathbb{E}_{q(\mathbf{x}_{0:t})} \Big[ \log \frac{q(\mathbf{x}_{1:t}\vert\mathbf{x}_0)}{p_\theta(\mathbf{x}_{0:t})} \Big] \\
&= L_{VUB} \\
&= \bar{L}_t
\end{aligned}  
\end{equation}

Then follow the derivation of \cite{DBLP:conf/icml/Sohl-DicksteinW15}, we have

\begin{equation}
\begin{aligned}
\bar{L}_t
&= \mathbb{E}_{q(\mathbf{x}_{0:t})} \Big[ \log\frac{q(\mathbf{x}_{1:t}\vert\mathbf{x}_0)}{p_\theta(\mathbf{x}_{0:t})} \Big] \\
&= \mathbb{E}_q \Big[ \log\frac{\prod_{i=1}^t q(\mathbf{x}_i\vert\mathbf{x}_{i-1})}{ p_\theta(\mathbf{x}_t) \prod_{i=1}^t p_\theta(\mathbf{x}_{i-1} \vert\mathbf{x}_i) } \Big] \\
&= \mathbb{E}_q \Big[ -\log p_\theta(\mathbf{x}_t) + \sum_{i=1}^t \log \frac{q(\mathbf{x}_i\vert\mathbf{x}_{i-1})}{p_\theta(\mathbf{x}_{i-1} \vert\mathbf{x}_i)} \Big] \\
&= \mathbb{E}_q \Big[ -\log p_\theta(\mathbf{x}_t) + \sum_{i=2}^t \log \frac{q(\mathbf{x}_i\vert\mathbf{x}_{i-1})}{p_\theta(\mathbf{x}_{i-1} \vert\mathbf{x}_i)} + \log\frac{q(\mathbf{x}_1 \vert \mathbf{x}_0)}{p_\theta(\mathbf{x}_0 \vert \mathbf{x}_1)} \Big] \\
&= \mathbb{E}_q \Big[ -\log p_\theta(\mathbf{x}_t) + \sum_{i=2}^t \log \Big( \frac{q(\mathbf{x}_{i-1} \vert \mathbf{x}_i, \mathbf{x}_0)}{p_\theta(\mathbf{x}_{i-1} \vert\mathbf{x}_i)}\cdot \frac{q(\mathbf{x}_i \vert \mathbf{x}_0)}{q(\mathbf{x}_{i-1}\vert\mathbf{x}_0)} \Big) + \log \frac{q(\mathbf{x}_1 \vert \mathbf{x}_0)}{p_\theta(\mathbf{x}_0 \vert \mathbf{x}_1)} \Big] \\
&= \mathbb{E}_q \Big[ -\log p_\theta(\mathbf{x}_t) + \sum_{i=2}^t \log \frac{q(\mathbf{x}_{i-1} \vert \mathbf{x}_i, \mathbf{x}_0)}{p_\theta(\mathbf{x}_{i-1} \vert\mathbf{x}_i)} + \sum_{i=2}^t \log \frac{q(\mathbf{x}_i \vert \mathbf{x}_0)}{q(\mathbf{x}_{i-1} \vert \mathbf{x}_0)} + \log\frac{q(\mathbf{x}_1 \vert \mathbf{x}_0)}{p_\theta(\mathbf{x}_0 \vert \mathbf{x}_1)} \Big] \\
&= \mathbb{E}_q \Big[ -\log p_\theta(\mathbf{x}_t) + \sum_{i=2}^t \log \frac{q(\mathbf{x}_{i-1} \vert \mathbf{x}_i, \mathbf{x}_0)}{p_\theta(\mathbf{x}_{i-1} \vert\mathbf{x}_i)} + \log\frac{q(\mathbf{x}_t \vert \mathbf{x}_0)}{q(\mathbf{x}_1 \vert \mathbf{x}_0)} + \log \frac{q(\mathbf{x}_1 \vert \mathbf{x}_0)}{p_\theta(\mathbf{x}_0 \vert \mathbf{x}_1)} \Big]\\
&= \mathbb{E}_q \Big[ \log\frac{q(\mathbf{x}_t \vert \mathbf{x}_0)}{p_\theta(\mathbf{x}_t)} + \sum_{i=2}^t \log \frac{q(\mathbf{x}_{i-1} \vert \mathbf{x}_i, \mathbf{x}_0)}{p_\theta(\mathbf{x}_{i-1} \vert\mathbf{x}_i)} - \log p_\theta(\mathbf{x}_0 \vert \mathbf{x}_1) \Big] \\
&= \mathbb{E}_q [\underbrace{D_\text{KL}(q(\mathbf{x}_t \vert \mathbf{x}_0) \parallel p_\theta(\mathbf{x}_t))}_{L_t} + \sum_{i=2}^{t} \underbrace{D_\text{KL}(q(\mathbf{x}_{i-1} \vert \mathbf{x}_i, \mathbf{x}_0) \parallel p_\theta(\mathbf{x}_{i-1} \vert\mathbf{x}_i))}_{L_{i-1}} \underbrace{- \log p_\theta(\mathbf{x}_0 \vert \mathbf{x}_1)}_{L_0} ]
\end{aligned}
\end{equation}

For diffusion model, the $L_i$ term is parameterized to minimize the difference of these two distributions $q(\mathbf{x}_{t-1} \vert \mathbf{x}_t, \mathbf{x}_0)$ and $p_\theta(\mathbf{x}_{t-1} \vert\mathbf{x}_t)$. Since both distributions are gaussian distributions, the problem can be translated to minimize the difference of their mean, which can be further derived into the following formulation:

\begin{equation}
\begin{aligned}
L_t &= D_\text{KL}(q(\mathbf{x}_t \vert \mathbf{x}_0) \parallel p_\theta(\mathbf{x}_t)) = C_t\\
L_i &= \mathbb{E}_{\mathbf{x}_0, \boldsymbol{\epsilon}} \Big[\frac{ (1 - \alpha_i)^2 }{2 \alpha_i (1 - \bar{\alpha}_i) \| \boldsymbol{\Sigma}_i \|^2_2} \|\boldsymbol{\epsilon}_i - \boldsymbol{\epsilon}_\theta(\sqrt{\bar{\alpha}_i}\mathbf{x}_0 + \sqrt{1 - \bar{\alpha}_i}\boldsymbol{\epsilon}_i, i)\|^2 \Big], i\in[1,t-1]\\
L_0 &= - \log p_\theta(\mathbf{x}_0 \vert \mathbf{x}_1)
\end{aligned}
\label{Equation:LS}
\end{equation}

Consequently, $\bar{L}_t$ can be finally formulated as:

\begin{equation}
\begin{aligned}
\bar{L}_t 
&= C_t + \sum_{i=1}^{t-1} \Big[\frac{ (1 - \alpha_i)^2 }{2 \alpha_i (1 - \bar{\alpha}_i) \| \boldsymbol{\Sigma}_i \|^2_2} E_i \Big] + L_0 ,\\
\end{aligned}
\end{equation}

where $E_i = \|\boldsymbol{\epsilon}_i - \boldsymbol{\epsilon}_\theta(\sqrt{\bar{\alpha}_i}\mathbf{x}_0 + \sqrt{1 - \bar{\alpha}_i}\boldsymbol{\epsilon}_i, i)\|^2$.

\end{proof}

\section{Proof of \textbf{Theorem 1.} in Sec. 3.4}
\label{appendix:A2}

\begin{proof} 
As shown in \KMEquation{Equation:LS}, when we change the initial $\mathbf{x}_0$ from high-resolution image $\mathbf{x}_0$ to low-resolution image $\hat{\mathbf{x}}_0$, the main different come from the $L_t$ term. Since $p_\theta(\mathbf{x}_t)$ does not change, we will focus on the change of $q(\mathbf{x}_t|\mathbf{x}_0)$. Note that $q(\mathbf{x}_t|\mathbf{x}_0)$ is the forward distribution which we have:

\begin{equation}
q(\mathbf{x}_t \vert \mathbf{x}_0) = \mathcal{N}(\mathbf{x}_t; \sqrt{\bar{\alpha}_t} \mathbf{x}_0, (1 - \bar{\alpha}_t)\mathbf{I})
\end{equation}

To analyze the changes, we can analyze the DK-divergence of these two distributions $q(\mathbf{x}_t|\mathbf{x}_0)$ and $q(\mathbf{x}_t|\hat{\mathbf{x}}_0)$.

\begin{equation}
\begin{aligned}
&D_\text{KL}(q(\mathbf{x}_t \vert \mathbf{x}_0) \parallel q(\mathbf{x}_t \vert \hat{\mathbf{x}}_0)) \\
=& \int q(\mathbf{x}_t|\mathbf{x}_0) \ln(\frac{q(\mathbf{x}_t|\mathbf{x}_0)}{q(\mathbf{x}_t|\hat{\mathbf{x}}_0)}) d\mathbf{x}\\
=& \int q(\mathbf{x}_t|\mathbf{x}_0) \ln(\frac{e^{-\frac{ (\mathbf{x}-\sqrt{\bar{\alpha}_t} \mathbf{x}_0)^2 }{(1-\bar{\alpha}_t)\mathbf{I}}}}{e^{-\frac{ (\mathbf{x}-\sqrt{\bar{\alpha}_t} \hat{\mathbf{x}}_0)^2 }{(1-\bar{\alpha}_t)\mathbf{I}}}}) d\mathbf{x} \\
=& \int q(\mathbf{x}_t|\mathbf{x}_0) (-\frac{ (\mathbf{x}-\sqrt{\bar{\alpha}_t} \mathbf{x}_0)^2 }{(1-\bar{\alpha}_t)\mathbf{I}} + \frac{ (\mathbf{x}-\sqrt{\bar{\alpha}_t} \hat{\mathbf{x}}_0)^2 }{(1-\bar{\alpha}_t)\mathbf{I}}) d\mathbf{x} \\
=& \frac{1}{(1-\bar{\alpha}_t)\mathbf{I}} \int q(\mathbf{x}_t|\mathbf{x}_0) (\bar{\alpha}_t (\mathbf{x}_0)^2 + \bar{\alpha}_t (\hat{\mathbf{x}}_0)^2 + 2 \sqrt{\bar{\alpha}_t}(\mathbf{x}_0 - \hat{\mathbf{x}}_0) \mathbf{x}) d\mathbf{x} \\
=& \frac{1}{(1-\bar{\alpha}_t)\mathbf{I}} \int q(\mathbf{x}_t|\mathbf{x}_0) (\bar{\alpha}_t (\mathbf{x}_0)^2 + \bar{\alpha}_t (\hat{\mathbf{x}}_0)^2) d\mathbf{x} \\
&\quad + \frac{1}{(1-\bar{\alpha}_t)\mathbf{I}} \int q(\mathbf{x}_t|\mathbf{x}_0) 2 \sqrt{\bar{\alpha}_t}(\mathbf{x}_0 - \hat{\mathbf{x}}_0) \mathbf{x}) d\mathbf{x} \\ 
=& \frac{\bar{\alpha}_t (\mathbf{x}_0)^2 + \bar{\alpha}_t (\hat{\mathbf{x}}_0)^2}{(1-\bar{\alpha}_t)\mathbf{I}} \int q(\mathbf{x}_t|\mathbf{x}_0) d\mathbf{x} + \frac{2 \sqrt{\bar{\alpha}_t}(\mathbf{x}_0 - \hat{\mathbf{x}}_0)}{(1-\bar{\alpha}_t)\mathbf{I}} \int q(\mathbf{x}_t|\mathbf{x}_0) \mathbf{x} d\mathbf{x} \\
=& \frac{\bar{\alpha}_t (\mathbf{x}_0)^2 + \bar{\alpha}_t (\hat{\mathbf{x}}_0)^2}{(1-\bar{\alpha}_t)\mathbf{I}} + \frac{2 \sqrt{\bar{\alpha}_t}(\mathbf{x}_0 - \hat{\mathbf{x}}_0)}{(1-\bar{\alpha}_t)\mathbf{I}} \int q(\mathbf{x}_t|\mathbf{x}_0) \mathbf{x} d\mathbf{x} \\
\end{aligned}
\end{equation}

As $q(\mathbf{x}_t|\mathbf{x}_0)$ is a gaussian distribution, $\int q(\mathbf{x}_t|\mathbf{x}_0) \mathbf{x} d\mathbf{x}$ is exactly the mean of this distribution which is $\sqrt{\bar{\alpha}_t}\mathbf{x}_0$, so we have

\begin{equation}
\begin{aligned}
D_\text{KL}(q(\mathbf{x}_t \vert \mathbf{x}_0) \parallel q(\mathbf{x}_t \vert \hat{\mathbf{x}}_0)) &= \frac{\bar{\alpha}_t (\mathbf{x}_0)^2 + \bar{\alpha}_t (\hat{\mathbf{x}}_0)^2}{(1-\bar{\alpha}_t)\mathbf{I}} + \frac{2 \sqrt{\bar{\alpha}_t}(\mathbf{x}_0 - \hat{\mathbf{x}}_0)}{(1-\bar{\alpha}_t)\mathbf{I}} \sqrt{\bar{\alpha}_t}\mathbf{x}_0 \\
\end{aligned}
\label{Equation:KL_FW}
\end{equation}

When we increase the noise injection level $s$, $\bar{\alpha}_t$ decreases, then we can find $D_\text{KL}(q(\mathbf{x}_t \vert \mathbf{x}_0) \parallel q(\mathbf{x}_t \vert \hat{\mathbf{x}}_0))$ decreases which means $q(\mathbf{x}_t \vert \hat{\mathbf{x}}_0)$ becomes more similar with $q(\mathbf{x}_t \vert \mathbf{x}_0)$. Especially, when $t\to \infty$, $\bar{\alpha}_t \to 0$, then $D_\text{KL}(q(\mathbf{x}_t \vert \mathbf{x}_0 \parallel q(\mathbf{x}_t \vert \hat{\mathbf{x}}_0))=0$, both $q(\mathbf{x}_t \vert \mathbf{x}_0)$ and $q(\mathbf{x}_t \vert \hat{\mathbf{x}}_0)$ follows the same gaussian distribution. Besides, we can also find the gap between these two distributions is linear to the error between the error $(\mathbf{x}_0-\hat{\mathbf{x}}_0)$ and decreases along with the noise injection level $t$. Then the $L_t$ term with $\hat{\mathbf{x}}_0$ as input can be:

\begin{equation}
\begin{aligned}
L_t &= D_\text{KL}(q(\mathbf{x}_t \vert \hat{\mathbf{x}}_0) \parallel p_\theta(\mathbf{x}_t))\\
& \triangleq D_\text{KL}(q(\mathbf{x}_t \vert \mathbf{x}_0)\parallel p_\theta(\mathbf{x}_t)) + D_\text{KL}(q(\mathbf{x}_t \vert \hat{\mathbf{x}}_0) \parallel q(\mathbf{x}_t \vert \mathbf{x}_0)) \\
&= L_t + A_t + K_t (\mathbf{x}_0-\hat{\mathbf{x}}_0) 
\end{aligned}
\end{equation}

Where $L_t$ is the forward error when we use $\mathbf{x}$ as input, $A_t$ is the first term in \KMEquation{Equation:KL_FW}, $K_t (\mathbf{x}_0-\hat{\mathbf{x}}_0) $ is the second term. The second line is because we have proved that the KL-divergence between $q(\mathbf{x}_t \vert \mathbf{x}_0)$ and $q(\mathbf{x}_t \vert \hat{\mathbf{x}}_0)$ becomes smaller as $t$ increase, so these two distributions become more and more similar as $t$ increases. So we can use $q(\mathbf{x}_t \vert \hat{\mathbf{x}}_0))$ to approximate $q(\mathbf{x}_t \vert \mathbf{x}_0)$ to some extent. Then we have:

\begin{equation}
\begin{aligned}
&D_\text{KL}(q(\mathbf{x}_t \vert \mathbf{x}_0)\parallel p_\theta(\mathbf{x}_t)) + D_\text{KL}(q(\mathbf{x}_t \vert \hat{\mathbf{x}}_0) \parallel q(\mathbf{x}_t \vert \mathbf{x}_0)) \\
=& \int q(\mathbf{x}_t \vert \mathbf{x}_0) \log \frac{q(\mathbf{x}_t \vert \mathbf{x}_0)}{p_\theta(\mathbf{x}_t)} + \int q(\mathbf{x}_t \vert \hat{\mathbf{x}}_0) \log \frac{q(\mathbf{x}_t \vert \hat{\mathbf{x}}_0)}{q(\mathbf{x}_t \vert \mathbf{x}_0)} \\
\triangleq & \int q(\mathbf{x}_t \vert \hat{\mathbf{x}}_0) \log \frac{q(\mathbf{x}_t \vert \mathbf{x}_0)}{p_\theta(\mathbf{x}_t)} + \int q(\mathbf{x}_t \vert \hat{\mathbf{x}}_0) \log \frac{q(\mathbf{x}_t \vert \hat{\mathbf{x}}_0)}{q(\mathbf{x}_t \vert \mathbf{x}_0)} \\
=& \int q(\mathbf{x}_t \vert \hat{\mathbf{x}}_0) \big( \log \frac{q(\mathbf{x}_t \vert \mathbf{x}_0)}{p_\theta(\mathbf{x}_t)} +  \log \frac{q(\mathbf{x}_t \vert \hat{\mathbf{x}}_0)}{q(\mathbf{x}_t \vert \mathbf{x}_0)} \big) \\
=& \int q(\mathbf{x}_t \vert \hat{\mathbf{x}}_0) \log \frac{q(\mathbf{x}_t \vert \mathbf{x}_0)}{p_\theta(\mathbf{x}_t)}
\end{aligned}
\end{equation}

Consequently, the final recovery error $\mathcal{L}_t$ can be revised as:

\begin{equation}
\begin{aligned}
\mathcal{L}_t &= L_t + \sum_{i=0}^{t-1} L_{i} \\
&= D_\text{KL}(q(\mathbf{x}_t \vert \hat{\mathbf{x}}_0) \parallel p_\theta(\mathbf{x}_t)) + \sum_{i=0}^{t-1} L_{i} \\
&\triangleq D_\text{KL}(q(\mathbf{x}_t \vert \mathbf{x}_0)\parallel p_\theta(\mathbf{x}_t)) + D_\text{KL}(q(\mathbf{x}_t \vert \hat{\mathbf{x}}_0) \parallel q(\mathbf{x}_t \vert \mathbf{x}_0)) + \sum_{i=0}^{t-1} L_{i}\\
&= \sum_{i=0}^{t} L_{i} + \big[ K_t (\mathbf{x}_0-\hat{\mathbf{x}}_0)  + A_t \big] \\
&\triangleq \bar{L}_t + \omega \big[ K_t \parallel \mathbf{x} - \hat{\mathbf{x}}_0 \parallel^2 + A_t \big] \\
&\triangleq \underbrace{\mathcal{L}^{S}_t}_{\text{\zqh{Signature Loss}}} +  
    \underbrace{ \omega \mathcal{L}^{F}_t}_{\zqh{L_1~\text{Fidelity Loss}}}
\end{aligned}
\end{equation}

where $\omega$ is a hyperparameter to guarantee these two loss terms are of the same magnitude, we choose $\omega=0.004$.

\end{proof}

\section{Details of Experimental Setup in Sec. 4.1}
\label{addition:details}

\noindent\textbf{Baselines and Datasets.} 
We use Bicubic \cite{Bicubic,Bicubic2}, Nearest \cite{Nearest}, EDSR \cite{EDSR}, ESRGAN \cite{ESRGAN}, LIIF \cite{LIIF}, and SR3 \cite{DBLP:journals/corr/abs-2104-07636} as the baseline solutions for performance comparison.

To guarantee evaluation fairness, the models of baseline solutions, as well as the pre-trained DGM used by Diff-SR, are established  on a unified experimental setup.
We use four pertinent SR datasets, including UDM10 \cite{udm10}, REDS \cite{reds}, VID \cite{VID}, and Vimeo90K \cite{vimeo}. 
Following the pre-training guidances mentioned in the corresponding work of the baseline solutions, we use the high-resolution (HR) images as the supervised ground truth to pre-train the baseline models, so that they can recover the low-resolution (LR) images to the HR versions.
Considering the property of the diffusion process, the DGM only requires the HR images to optimize its restoration capacity, without access need for the LR images.

\noindent\textbf{Setting of Pre-trained DGM.} 
The DGM used by Diff-SR employs the UNet backbone \cite{Unet} and follows the sampling strategy of \textit{Denoising Diffusion Implicit Models} (DDIMs) \cite{DBLP:conf/iclr/SongME21}.
More precisely, the DGM uses three downsampling blocks, two middle blocks and three upsampling blocks in UNet. The scaling factors are set as $2$, $4$, $8$ for these three kinds of blocks, respectively. The number of base feature channels is $64$. 
To capture the time sequence information, the diffusion step index $t$ is specified by adding the sinusoidal position embedding into each residual block. 
By setting the maximum step number as $T=1000$, the DGM controls the noise variance $\beta_t$ ($t \in [1, T]$) through a linear quadratic scheduler, which gradually ranges from $\beta_1=10^{-4}$ to $\beta_T=0.02$. 
Also, the DGM is optimized by the \textit{Mean of Squared Error} (MSE) loss with Adam optimizer \cite{Adam} and $16$ batch size. The total number of epochs is $10$K and the initial learning rate is $1 \times 10^{-5}$.

\noindent\textbf{Test setting.} 
Our Diff-SR and the other baseline solutions take the LR images to generate the recovered HR images.
The performance is compared by checking the scores of FID, SSIM and PSNR, between the recovered HR images and the ground-truth HR ones.

\section{Additional Performance Comparison in Sec. 4.2}
\label{Section:other_tables}

In \KMTable{Table:evaluation_VID} and \KMTable{Table:evaluation_vimeo}, we demonstrate the additional performance comparison between our Diff-SR and the baseline solutions, on the datasets of VID \cite{VID} and Vimeo90K \cite{vimeo}. It is clear that our Diff-SR consistently achieves superior performance over the baseline solutions, in terms of FID, PSNR and SSIM.

\begin{table*}[ht]
\begin{center}
\caption{Comparison with baseline solutions released by OpenMMLab on VID dataset, where the \redContent{\textbf{red}} and
\blueContent{blue} colors indicate the best and the second-best performance, respectively. } 
\label{Table:evaluation_VID} 
\resizebox{\linewidth}{!}{
\begin{tabular}{*{14}{c}}
\toprule
\multirow{2}*{\makecell[c]{Model}} & \multicolumn{3}{c}{\makecell[c]{$2\times$}} & \multicolumn{3}{c}{\makecell[c]{$2.7\times$}} & \multicolumn{3}{c}{\makecell[c]{$3.5\times$}} & \multicolumn{3}{c}{\makecell[c]{$4\times$}} &\\
\cmidrule(lr){2-4}\cmidrule(lr){5-7}\cmidrule(lr){8-10}\cmidrule(lr){11-13}
& FID $\downarrow$ & PSNR $\uparrow$ & SSIM $\uparrow$ & FID $\downarrow$ & PSNR $\uparrow$ & SSIM $\uparrow$ & FID $\downarrow$ & PSNR $\uparrow$ & SSIM $\uparrow$ & FID $\downarrow$ & PSNR $\uparrow$ & SSIM $\uparrow$ \\
\midrule
Nearest & 6.696 & 30.639 & 0.757 & 8.670 & 30.111 & 0.641 & 12.594 & 29.684 & 0.534 & 14.583 & 29.529 & 0.492 \\
Bicubic & 11.549 & 30.758 & 0.777 & 18.134 & 30.208 & 0.681 & 28.006 & 29.717 & 0.573 & 32.888 & 29.531 & 0.529 \\
EDSR + Bicubic & \blueContent{\textbf{0.302}} & 30.204 & 0.773 & 5.798 & 28.794 & 0.496 & 3.661 & 28.712 & 0.429 & 3.761 & 28.669 & 0.388 \\
ESRGAN + Bicubic & 81.632 & 27.862 & 0.187 & 74.579 & 27.802 & 0.145 & 77.155 & 27.801 & 0.144 & 73.057 & 27.862 & 0.160 \\
LIIF & 4.289 & \blueContent{\textbf{31.429}} & 0.860 & 6.084 & \blueContent{\textbf{31.094}} & \blueContent{\textbf{0.820}} & 10.889 & \blueContent{\textbf{30.430}} & \blueContent{\textbf{0.734}} & 11.943 & 30.004 & 0.661 \\
SR3 + Bicubic & 1.189 & 27.793 & \blueContent{\textbf{0.952}} & \blueContent{\textbf{0.645}} & 29.000 & 0.683 &\blueContent{\textbf{ 0.798}} & 28.916 & 0.649 & \blueContent{\textbf{0.322}} & \blueContent{\textbf{30.339}} & \redContent{\textbf{0.977}} \\
\midrule
\textbf{Diff-SR} & \redContent{\textbf{0.047}} & \redContent{\textbf{34.191}} & \redContent{\textbf{0.953}} & \redContent{\textbf{0.054}} & \redContent{\textbf{34.497}} & \redContent{\textbf{0.957}} & \redContent{\textbf{0.066}} & \redContent{\textbf{33.867}} & \redContent{\textbf{0.943}} & \redContent{\textbf{0.052}} & \redContent{\textbf{32.903}} & \blueContent{\textbf{0.913}} \\
\bottomrule
\end{tabular}
}
\end{center}
\end{table*}

\begin{table*}[ht]
\begin{center}
\caption{Comparison with baseline solutions released by OpenMMLab on VIMEO90K dataset, where the \redContent{\textbf{red}} and
\blueContent{blue} colors indicate the best and the second-best performance, respectively.} 
\label{Table:evaluation_vimeo} 
\resizebox{\linewidth}{!}{
\begin{tabular}{*{14}{c}}
\toprule
\multirow{2}*{\makecell[c]{Model}} & \multicolumn{3}{c}{\makecell[c]{$2\times$}} & \multicolumn{3}{c}{\makecell[c]{$2.7\times$}} & \multicolumn{3}{c}{\makecell[c]{$3.5\times$}} & \multicolumn{3}{c}{\makecell[c]{$4\times$}} &\\
\cmidrule(lr){2-4}\cmidrule(lr){5-7}\cmidrule(lr){8-10}\cmidrule(lr){11-13}
& FID $\downarrow$ & PSNR $\uparrow$ & SSIM $\uparrow$ & FID $\downarrow$ & PSNR $\uparrow$ & SSIM $\uparrow$ & FID $\downarrow$ & PSNR $\uparrow$ & SSIM $\uparrow$ & FID $\downarrow$ & PSNR $\uparrow$ & SSIM $\uparrow$ \\
\midrule
Nearest & 2.320 & 34.472 & 0.886 & 2.792 & 33.521 & 0.827 & 3.740 & 32.712 & 0.762 & 4.145 & 32.432 & 0.736 \\
Bicubic & 3.486 & \blueContent{\textbf{35.165}} & \blueContent{\textbf{0.907}} & 5.508 & 34.163 & 0.863 & 8.553 & 33.189 & 0.807 & 10.015 & 32.824 & 0.782 \\
EDSR + Bicubic & 1.214 & 32.898 & 0.844 & 8.165 & 30.835 & 0.659 & 10.066 & 30.269 & 0.569 & 13.515 & 29.748 & 0.476 \\
ESRGAN + Bicubic & \redContent{\textbf{0.605}} & 32.120 & 0.843 & \blueContent{\textbf{0.589}} & 31.440 & 0.857 & \blueContent{\textbf{0.962}} & 31.054 & 0.800 & \blueContent{\textbf{1.023}} & 30.716 & 0.744 \\
LIIF & 1.734 & \redContent{\textbf{36.456}} & \redContent{\textbf{0.942}} & 2.911 & \blueContent{\textbf{35.689}} & \blueContent{\textbf{0.919}} & 5.193 & \blueContent{\textbf{34.459}} & \blueContent{\textbf{0.877}} & 6.745 & \redContent{\textbf{33.643}} & \redContent{\textbf{0.838}} \\
SR3 + Bicubic & 1.287 & 27.777 & 0.817 & 13.601 & 27.970 & 0.490 & 15.617 & 27.968 & 0.459 & \redContent{\textbf{0.817}} & 28.811 & 0.803 \\
\midrule
\textbf{Diff-SR} & \blueContent{\textbf{1.112}} & 34.252 & 0.875 & \redContent{\textbf{0.517}} & \blueContent{\textbf{35.362}} & \redContent{\textbf{0.922}} & \redContent{\textbf{0.580}} & \redContent{\textbf{35.001}} & \redContent{\textbf{0.916}} & 1.122 & \blueContent{\textbf{33.180}} & \blueContent{\textbf{0.834}} \\
\bottomrule
\end{tabular}
}
\end{center}
\end{table*}

\clearpage

\section{Deep Inspection from Frequency Domain}
\label{section:LF_HF}

\begin{figure}[htp]
  \centering
    \includegraphics[width=\linewidth]{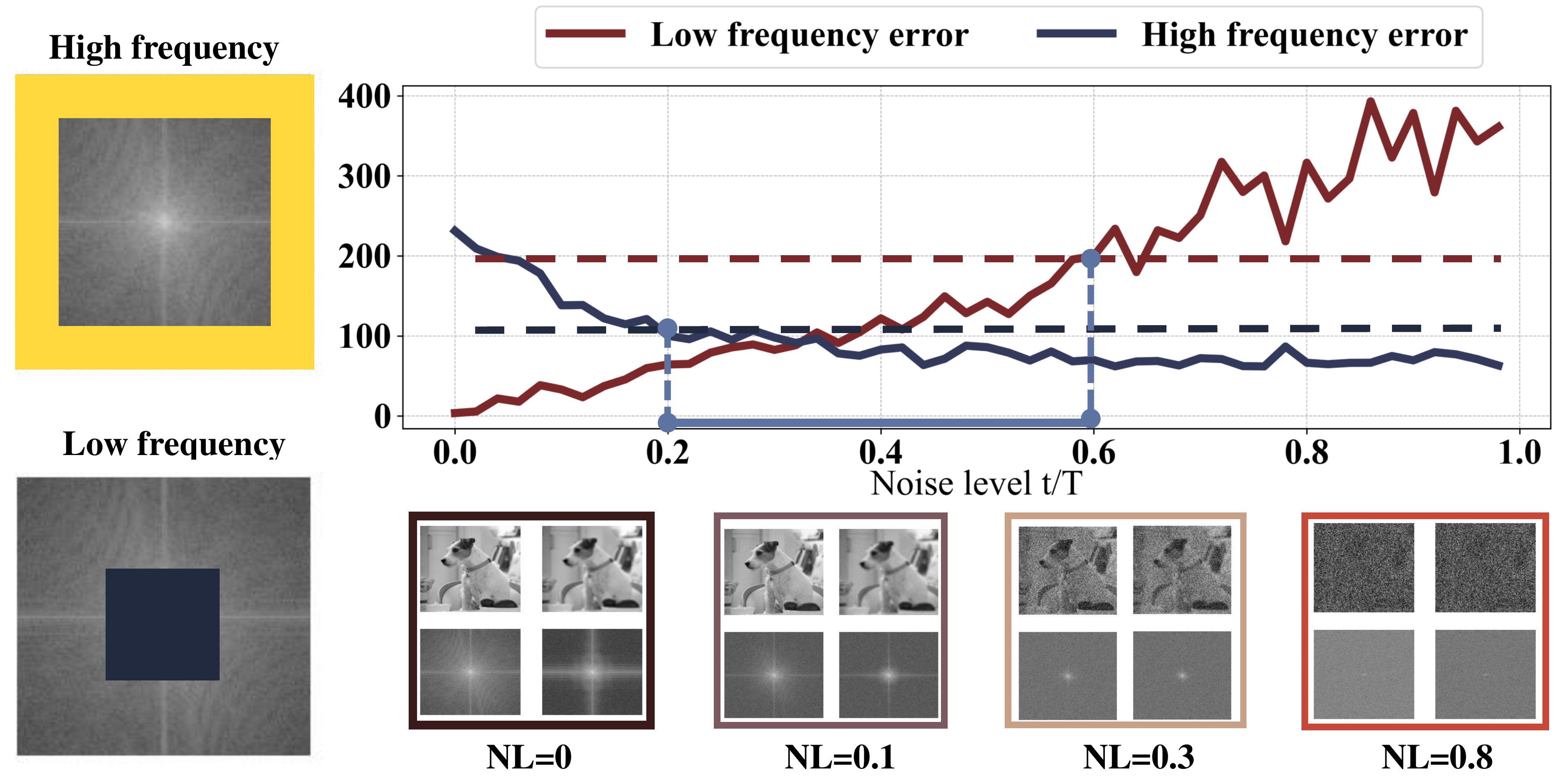}
  \caption{Illustration on frequency domain. For each box in the figure, the upper left is the result of the \textbf{original image} inject $t$ steps noise, the lower left is the frequency map of this image, the upper right is the result of \textbf{low-resolution image} inject $t$ steps noise, lower right is the frequency map.}
  \label{Figure:Frequence_domain}
\end{figure}

Another interesting result is when we analyze this problem in the frequency domain (through Fourier transform \cite{Fourier,Fourier2}), it gives us another perspective to understand why this noise injection operation can restore a low-resolution image to a high-resolution one. As shown in \KMFigure{Figure:Frequence_domain}, we calculate the model output $\tilde{\mathbf{x}}$'s frequency map and compare it with the frequency map of the original image $\mathbf{x}$, we use two boxes to segment the low-frequency information and high-frequency information as shown in the supplementary material of \S \ref{section:LF_HF}. Besides, we also illustrate the frequency map of the intermediate results that we get after injecting noise, which is shown in the lower side of \KMFigure{Figure:Frequence_domain}. 

Different images in the spatial and frequency domains are shown in the first box at the initial state. Compared with the original image $\mathbf{x}_0$, the low-resolution image lost most of the high-frequency information (\ie four corners) and the low-frequency information is preserved. Then when we inject some noise into these images, their frequency maps also change. But as shown in the second box, if the noise injection level is insufficient, there are still some significant differences. As a result, when we compare the enhanced result $\tilde{\mathbf{x}},\tilde{\mathbf{x}}\sim q(\mathbf{x}|\hat{\mathbf{x}}_t)$ with ground truth $\mathbf{x}_0$, we can find the low-frequency error is small but the high-frequency error is large. From the spatial domain, this means the content of the output image is consistent with the ground truth, but the image is blurred. Then if we inject enough noise (\ie in the third box), the frequency domain maps are very similar. We can get a clear output image when we denoise this image with DGM. Both high-frequency and low-frequency errors are less than a threshold, so the output image is clear and content-right. If we inject too much noise as shown in the final box, the frequency maps are similar, but the low-frequency information is destroyed. Then when we start the reverse process at this data point, the output image suffers from high low-frequency error even though the low-frequency error is small. From the spatial domain, this means we have a clear output image but the semantic content of this image is different from the original image.

\section{Qualitative Results in Sec. 4.2}
\label{section:addition_images}

\begin{figure}[htp]
  \centering
    \includegraphics[width=\linewidth]{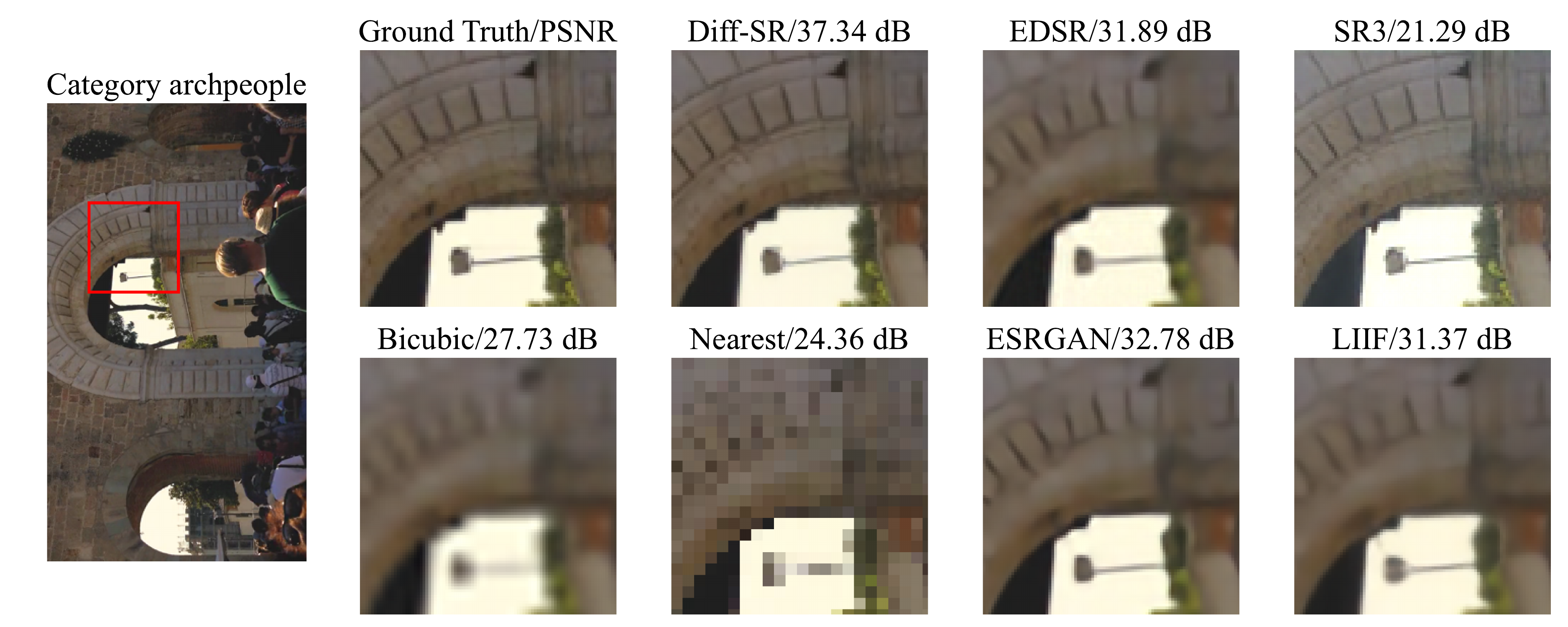}
  \caption{Qualitative results of different $4\times$ SR solutions, evaluated on UDM10 dataset.}
  \label{Figure:addition_udm10_1}
\vspace{-3pt}
\end{figure}

\begin{figure}[htp]
  \centering
    \includegraphics[width=\linewidth]{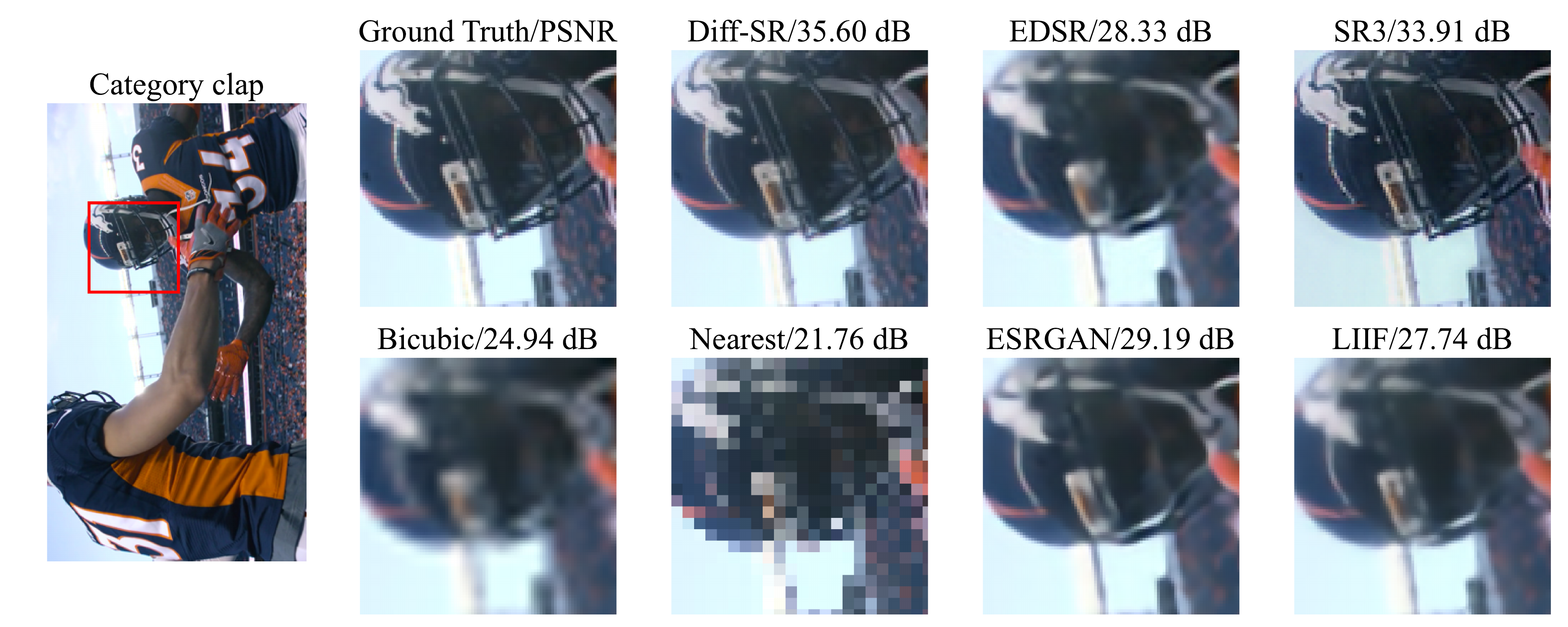}
  \caption{Qualitative results of different $4\times$ SR solutions, evaluated on UDM10 dataset.}
  \label{Figure:addition_udm10_2}
\vspace{-3pt}
\end{figure}

\begin{figure}[htp]
  \centering
    \includegraphics[width=\linewidth]{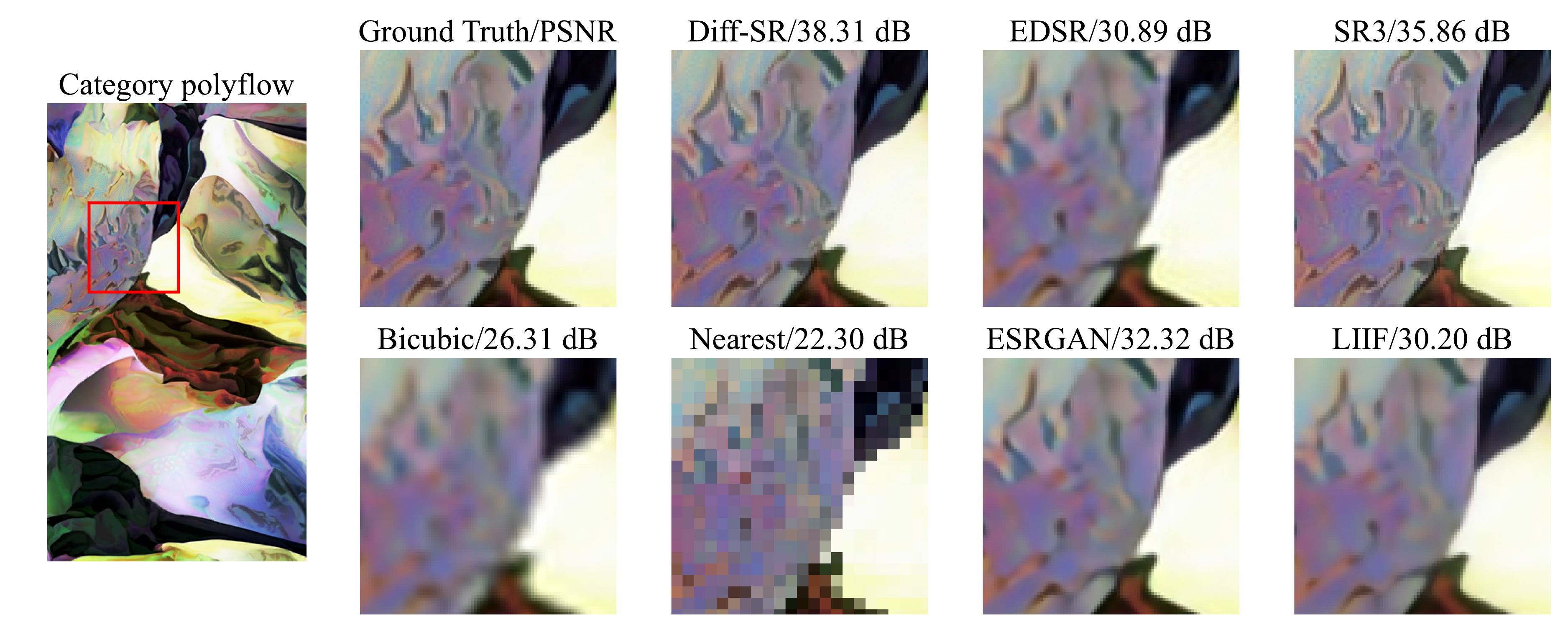}
  \caption{Qualitative results of different $4\times$ SR solutions, evaluated on UDM10 dataset.}
  \label{Figure:addition_udm10_3}
\end{figure}

\begin{figure}[htp]
  \centering
    \includegraphics[width=\linewidth]{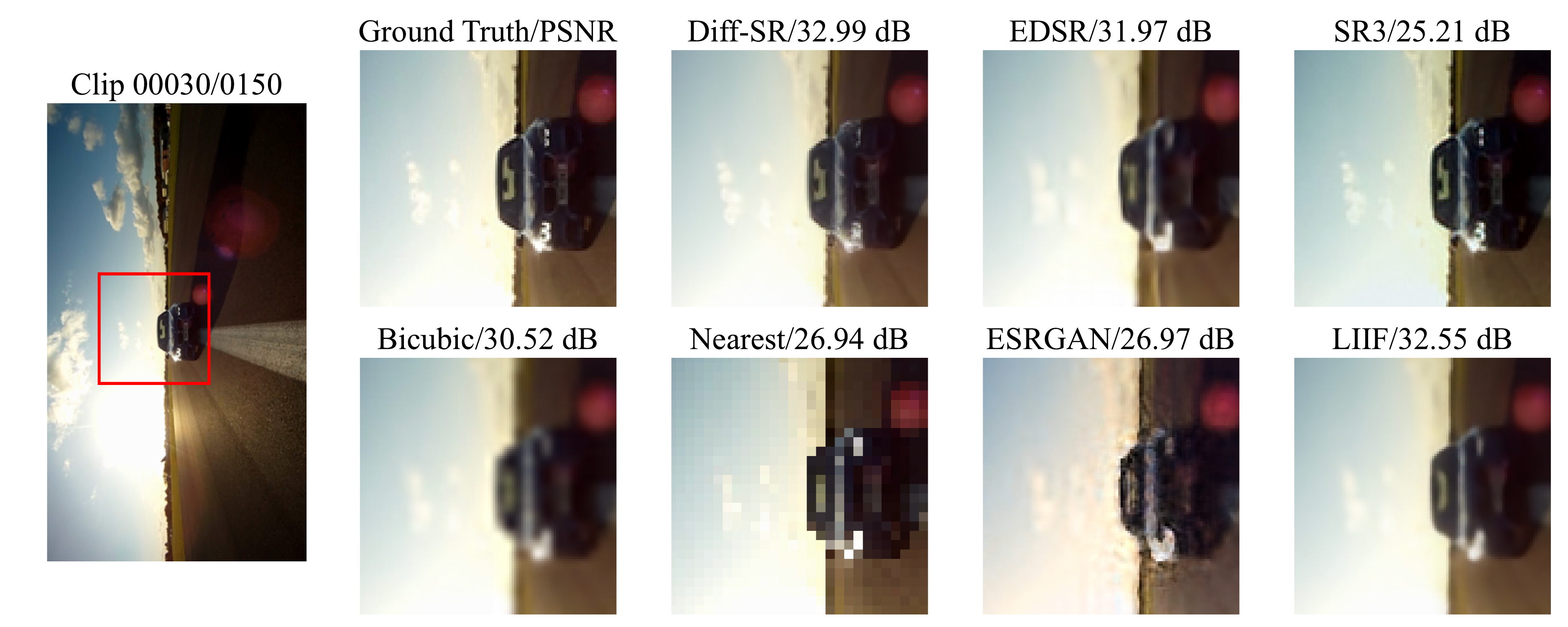}
  \caption{Qualitative results of different $4\times$ SR solutions, evaluated on VIMEO90K dataset.}
  \label{Figure:addition_vimeo_1}
\end{figure}

\begin{figure}[htp]
  \centering
    \includegraphics[width=\linewidth]{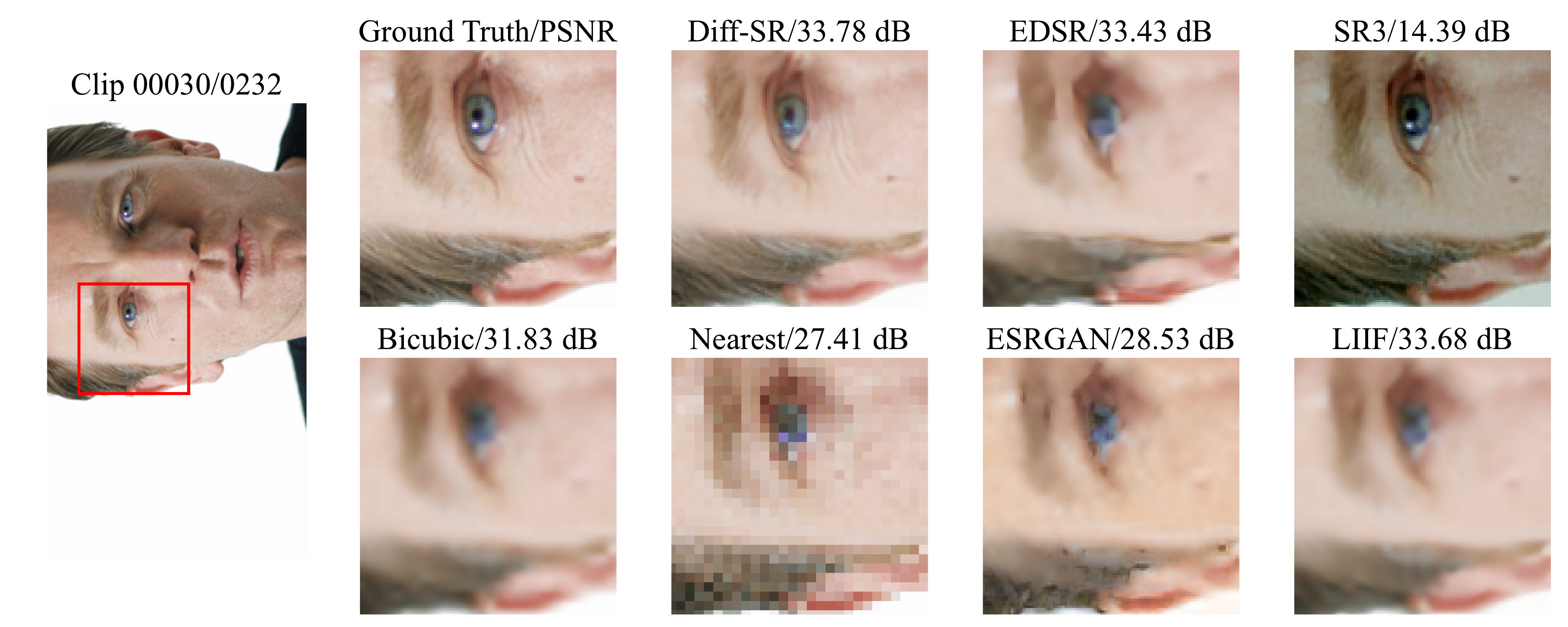}
  \caption{Qualitative results of different $4\times$ SR solutions, evaluated on VIMEO90K dataset.}
  \label{Figure:addition_vimeo_2}
\end{figure}

\begin{figure}[htp]
  \centering
    \includegraphics[width=\linewidth]{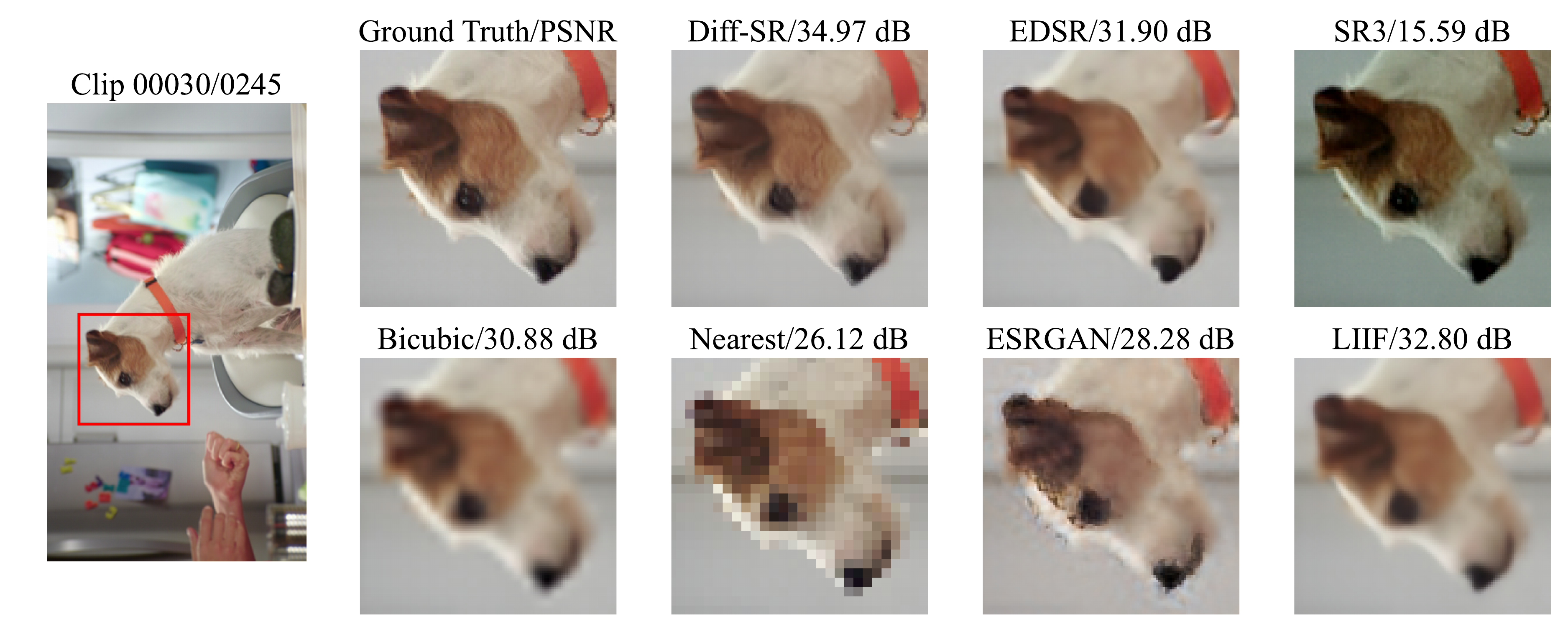}
  \caption{Qualitative results of different $4\times$ SR solutions, evaluated on VIMEO90K dataset.}
  \label{Figure:addition_vimeo_3}
\end{figure}

\begin{figure}[htp]
  \centering
    \includegraphics[width=\linewidth]{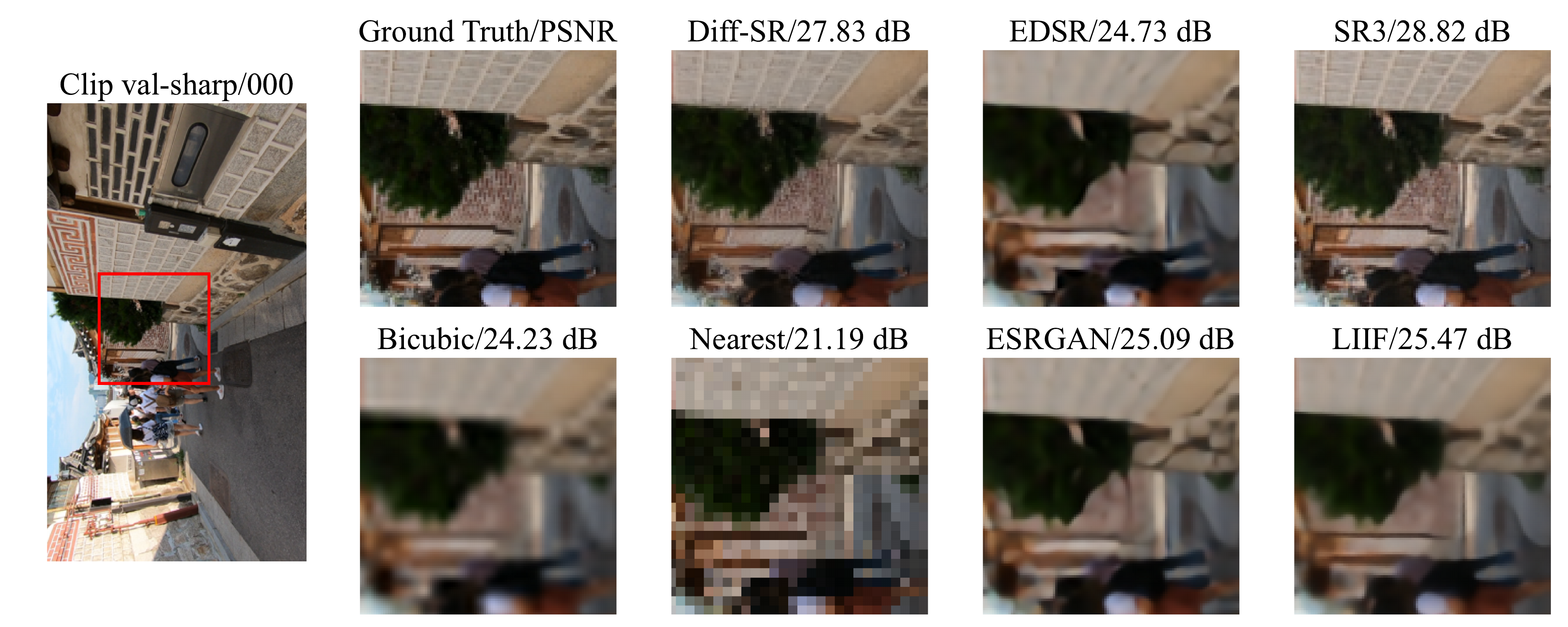}
  \caption{Qualitative results of different $4\times$ SR solutions, evaluated on REDS dataset.}
  \label{Figure:addition_reds_1}
\end{figure}

\begin{figure}[htp]
  \centering
    \includegraphics[width=\linewidth]{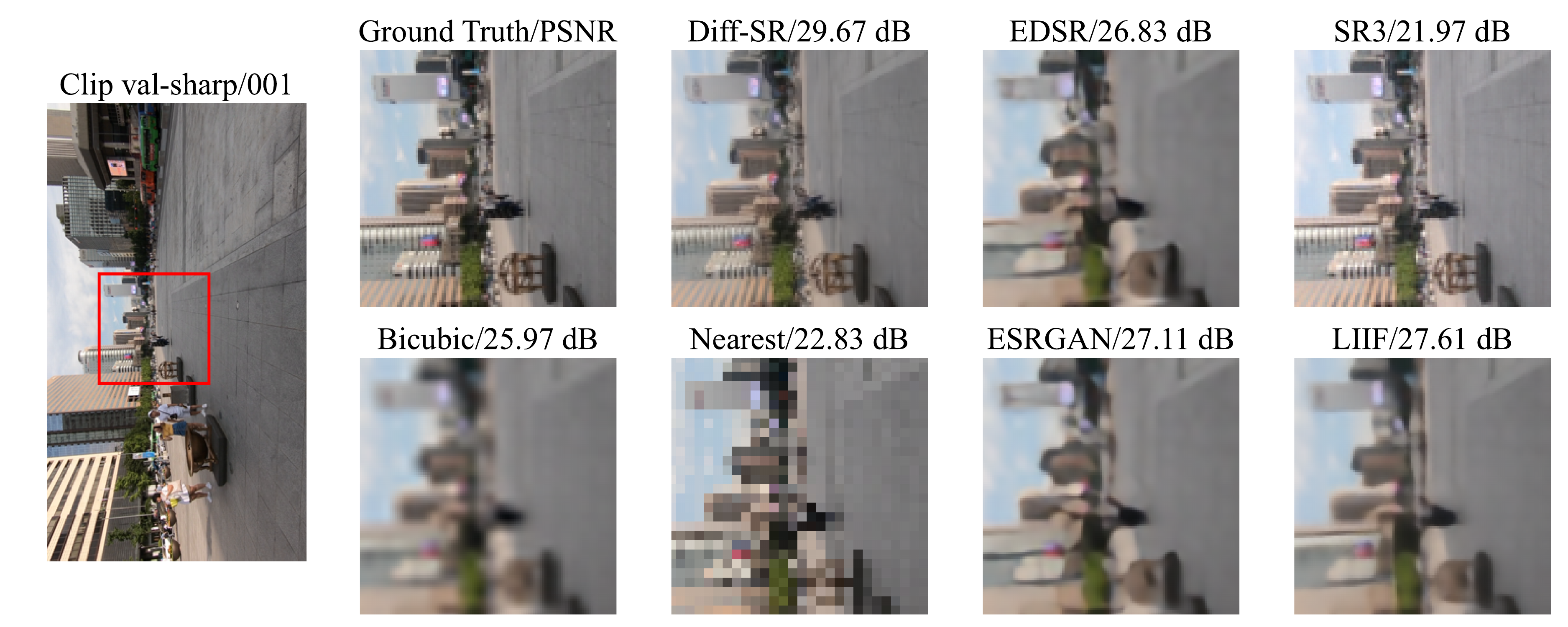}
  \caption{Qualitative results of different $4\times$ SR solutions, evaluated on REDS dataset.}
  \label{Figure:addition_reds_2}
\end{figure}

\begin{figure}[htp]
  \centering
    \includegraphics[width=\linewidth]{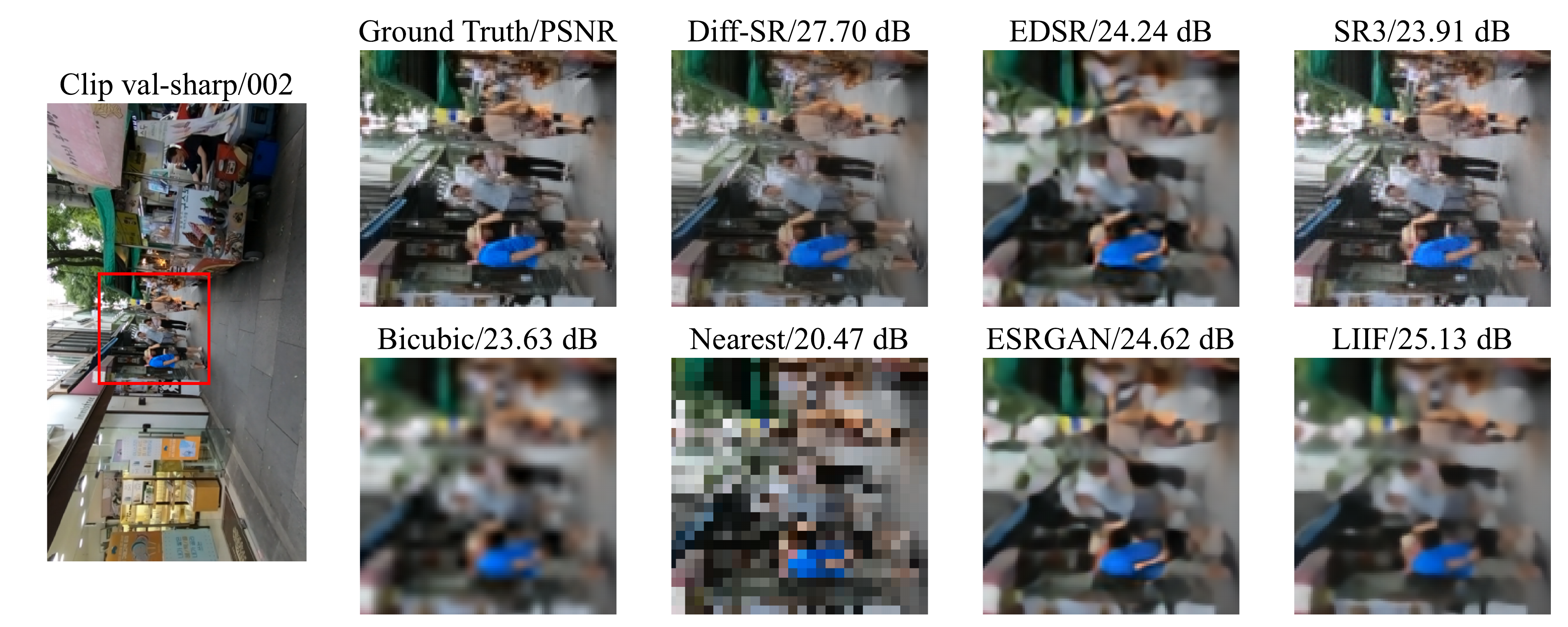}
  \caption{Qualitative results of different $4\times$ SR solutions, evaluated on REDS dataset.}
  \label{Figure:addition_reds_3}
\end{figure}

\begin{figure}[htp]
  \centering
    \includegraphics[width=\linewidth]
    {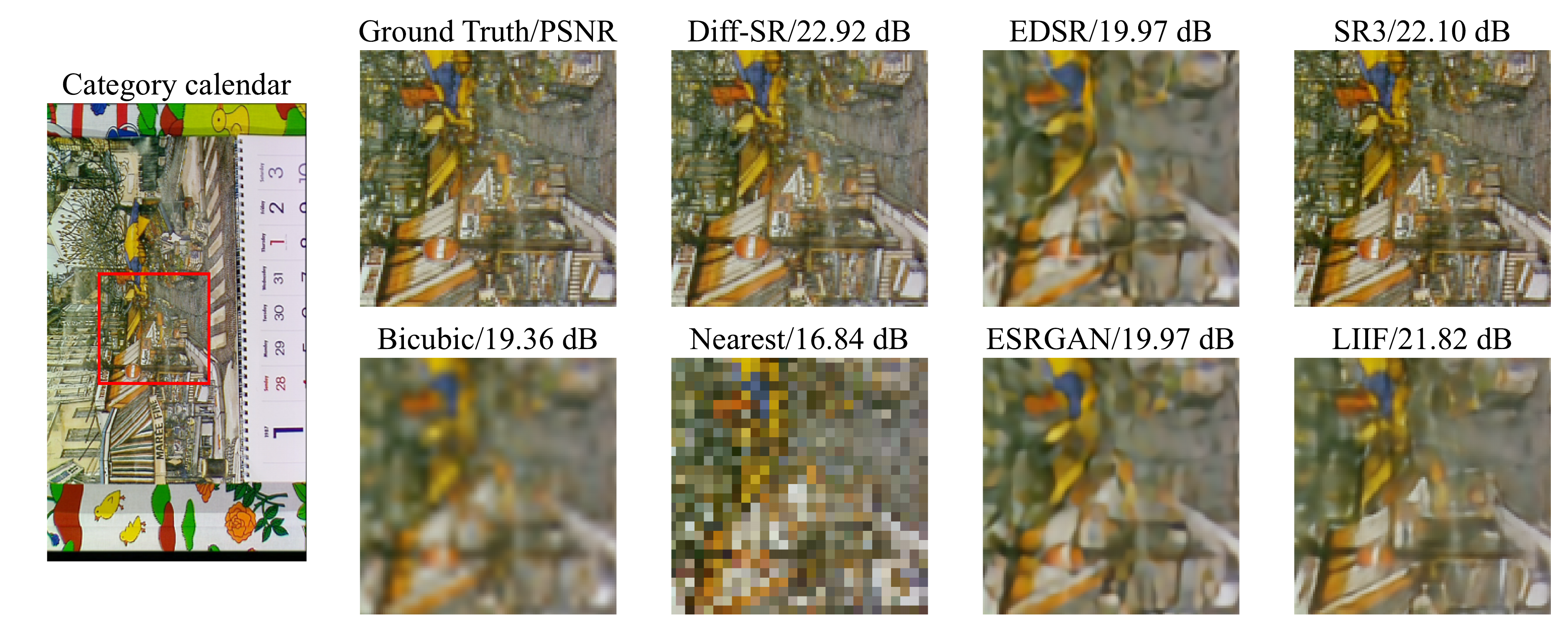}
  \caption{Qualitative results of different $4\times$ SR solutions, evaluated on VID4 dataset.}
  \label{Figure:addition_vid_1}
\end{figure}

\begin{figure}[htp]
  \centering
    \includegraphics[width=\linewidth]{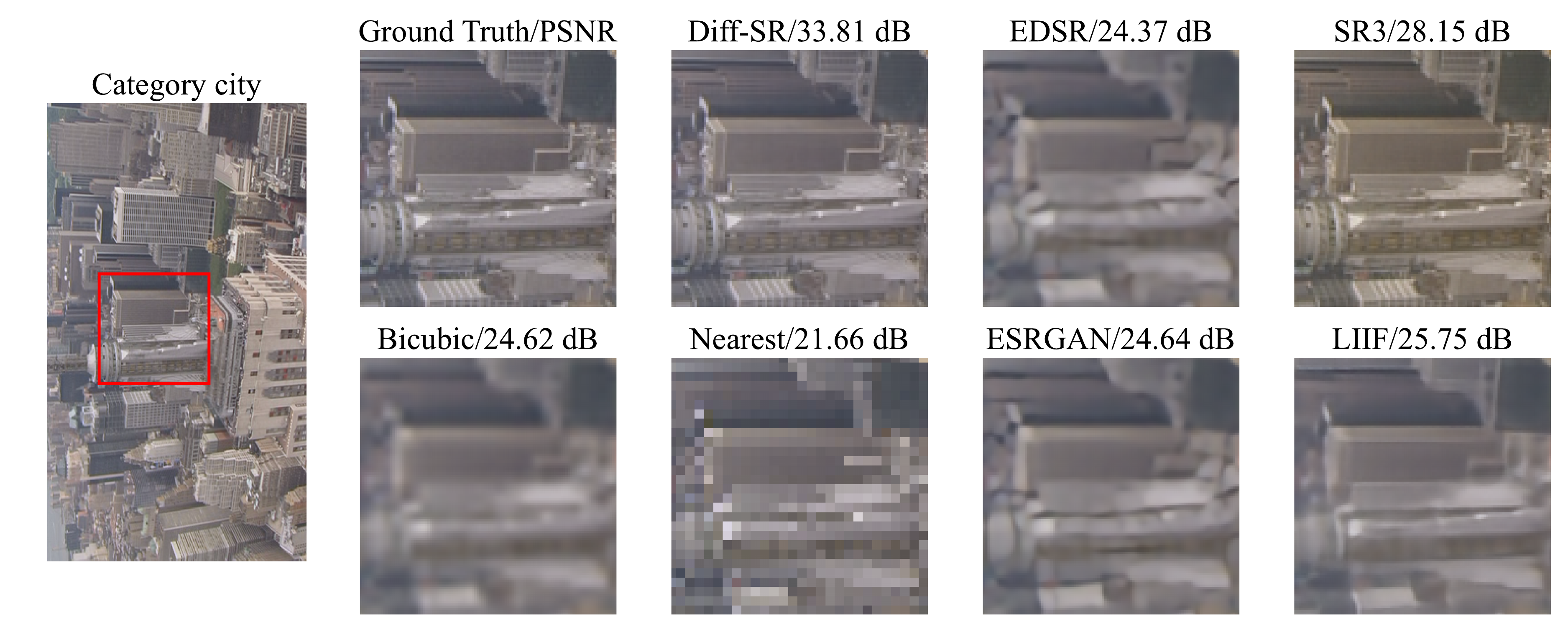}
  \caption{Qualitative results of different $4\times$ SR solutions, evaluated on VID4 dataset.}
  \label{Figure:addition_vid_2}
\end{figure}

\begin{figure}[htp]
  \centering
    \includegraphics[width=\linewidth]
    {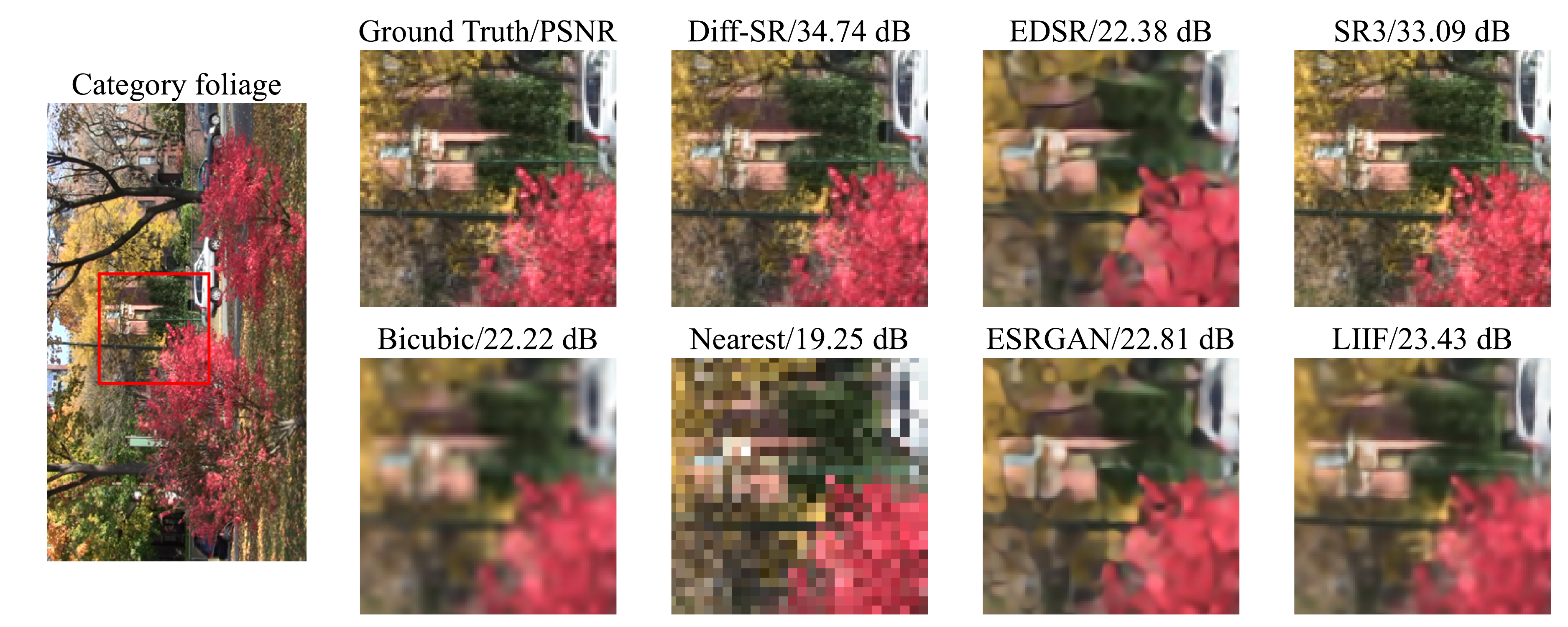}
  \caption{Qualitative results of different $4\times$ SR solutions, evaluated on VID4 dataset.}
  \label{Figure:addition_vid_3}
\end{figure}

\begin{figure}[htp]
  \centering
    \includegraphics[width=0.76\linewidth]{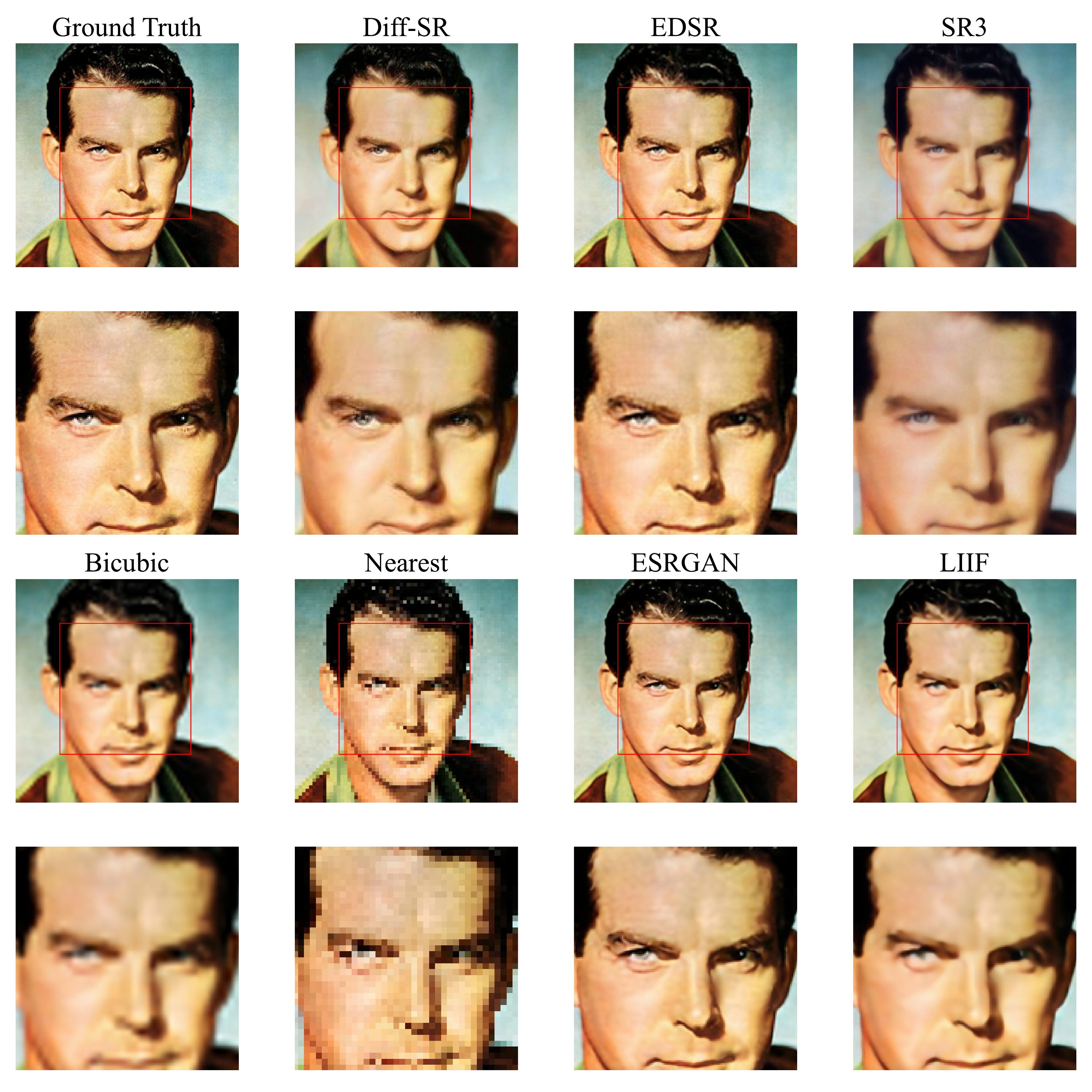}
  \caption{Qualitative results of different $4\times$ SR solutions, evaluated on CelebA dataset.}
  \label{Figure:addition_CelebA_1}
\vspace{-10pt}
\end{figure}

\begin{figure}[htp]
  \centering
    \includegraphics[width=0.76\linewidth]{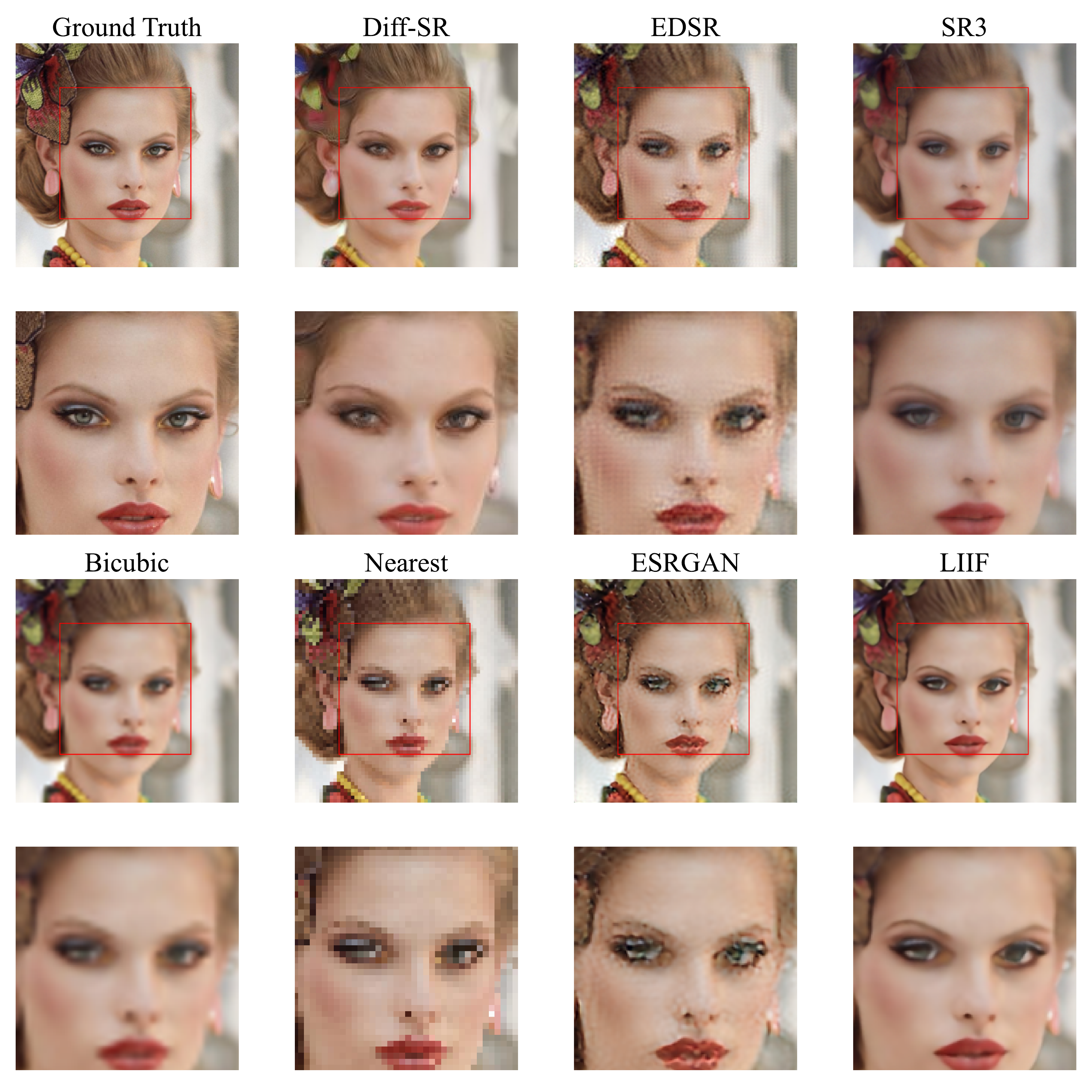}
  \caption{Qualitative results of different $4\times$ SR solutions, evaluated on CelebA dataset.}
  \label{Figure:addition_CelebA_2}
\end{figure}

In this section, we provide the qualitative results on the datasets of UDM10 \cite{udm10}, REDS \cite{reds}, VID \cite{VID}, and Vimeo90K \cite{vimeo}.
From \KMFigure{Figure:addition_udm10_1} to \KMFigure{Figure:addition_CelebA_2}, we can observe that Diff-SR consistently achieves higher recovery quality over the baseline solutions.

\clearpage

\end{document}